\documentclass{article}
\usepackage{jfrExamplee}
\usepackage{graphicx}
\usepackage{apalike}
\usepackage{setspace}

\usepackage{caption}
\usepackage[labelformat=simple]{subcaption}
\usepackage{booktabs}
\usepackage{multicol}
\usepackage{amsmath}
\usepackage{amsfonts}
\usepackage{amssymb}

\newcommand{\interval}[1]{[#1^{-}, #1^{+}]}
\newcommand{\asin}{\sin^{-1}}

\usepackage{cite}
\usepackage{hyperref}
\hypersetup{colorlinks,breaklinks,
    linkcolor=[rgb]{0.0,0.,0.},
    citecolor=[rgb]{0.000,0.0,0.0},
    urlcolor=[rgb]{0.031,0.318,0.612}}

\usepackage{acronym}
\acrodef{GESTALT}{Grid-based Estimation of Surface Traversability Applied to Local Terrain}
\acrodef{JPL}{Jet Propulsion Laboratory}
\acrodef{MSL}{Mars Science Laboratory}
\acrodef{MER}{Mars Exploration Rover}
\acrodef{M2020}{Mars 2020}
\acrodef{RSVP}{Rover Sequencing and Visualization Program}
\acrodef{ROAMS}{Rover Analysis Modeling and Simulation}
\acrodef{RBD}{Rocker-Bogie-Differential}
\acrodef{RD}{Rocker-Differential}
\acrodef{ACE}{Approximate Clearance Evaluation}
\acrodef{DEM}{Digital Elevation Map}
\acrodef{CFA}{Cumulative Fractional Area}
\acrodef{ODE}{Open Dynamics Engine}

\title{Fast Approximate Clearance Evaluation for Rovers with Articulated Suspension Systems}

\author{%
Kyohei Otsu\\
Jet Propulsion Laboratory\\
California Institute of Technology\\
Pasadena, CA 91109\\
\texttt{\{firstname.lastname\}@jpl.nasa.gov}\\
\And
Guillaume Matheron\\
\`{E}cole Normale Sup\`{e}rieure de Paris\\
Paris, France\\
\texttt{guillaume\_pub@matheron.eu}\\
\AND
Sourish Ghosh\\
Indian Institute of Technology\\
Kharagpur, India\\
\texttt{sourishg@iitkgp.ac.in}\\
\And
Olivier Toupet and  Masahiro Ono \\
Jet Propulsion Laboratory\\
California Institute of Technology\\
Pasadena, CA 91109 \\
\texttt{\{firstname.lastname\}@jpl.nasa.gov}
}

\begin{document}

\maketitle

\begin{abstract}
We present a light-weight body-terrain clearance evaluation algorithm for the automated path planning of NASA's Mars 2020 rover. Extraterrestrial path planning is challenging due to the combination of terrain roughness and severe limitation in computational resources. Path planning on cluttered and/or uneven terrains requires repeated safety checks on all the candidate paths at a small interval. Predicting the future rover state requires simulating the vehicle settling on the terrain, which involves an inverse-kinematics problem with iterative nonlinear optimization under geometric constraints. However, such expensive computation is intractable for slow spacecraft computers, such as RAD750, which is used by the Curiosity Mars rover and upcoming Mars 2020 rover. We propose the Approximate Clearance Evaluation (ACE) algorithm, which obtains conservative bounds on vehicle clearance, attitude, and suspension angles \textit{without} iterative computation. It obtains those bounds by estimating the lowest and highest heights that each wheel may reach given the underlying terrain, and calculating the worst-case vehicle configuration associated with those extreme wheel heights. The bounds are guaranteed to be conservative, hence ensuring vehicle safety during autonomous navigation. ACE is planned to be used as part of the new onboard path planner of the Mars 2020 rover. This paper describes the algorithm in detail and validates our claim of conservatism and fast computation through experiments.
\end{abstract}

\section{Introduction} \label{sec:intro}

\begin{figure}
    \centering
    \includegraphics[width=0.7\textwidth]{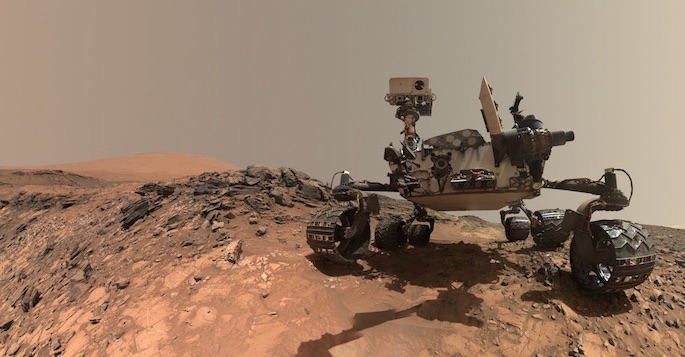}  
    \caption{Self-portrait of the MSL Curiosity rover, taken on Sol 1065. Sufficient clearance between the vehicle and the ground is necessary to safely traverse on rough terrain with outcrops. (Credit: NASA/JPL-Caltech/MSSS)}
    \label{fig:curiosity_selfie}
\end{figure}

Future planetary missions will require long-distance autonomous traverse on challenging, obstacle-rich terrains. For example, the landing site for the NASA/JPL \ac{M2020} mission will be the Jezero crater, a 49\,[km]-wide crater considered to be an ancient Martian lake produced by the past water-related activities \cite{Goudge2015}. 
Autonomous driving in this crater is expected to be challenging due to its high rock abundance. 
The state-of-the-art on-board path planner for Mars rovers called GESTALT \cite{Maimone2006}, which has successfully driven Spirit, Opportunity, and Curiosity rovers (\autoref{fig:curiosity_selfie}), is known to suffer from high rock density due to its highly conservative design. 
More specifically, GESTALT frequently fails to find a feasible path through a terrain  with 10\% \ac{CFA} (cumulative fraction of area covered by rocks), where CFA is a commonly used measure of rock abundance on Mars \cite{Golombek2008_CFA}. 
The Jezero crater has significantly higher rock density than any landing sites of previous Mars rover missions, where the CFA is up to 15--20\% based on the orbital reconaissance \cite{Golombek2015}. 
For this reason, a significant improvement in the autonomous driving capability was demanded by the Mars 2020 rover mission.

Conservatism is both a virtue and a limitation for spacecraft software. 
In general, any on-board algorithms must be conservative by design because no one can go to Mars to fix rovers if something goes wrong. 
In case of collision check in path planning, for example, false positives (a safe path is incorrectly assessed to be unsafe) are acceptable but false negatives (an unsafe path is incorrectly assessed to be safe) are unacceptable. 
However, excessive conservatism (i.e., too frequent false positives) results in reduced efficiency (e.g., unnecessarily winding paths) or inability to find solutions. 
Therefore, we have two adversarial objectives: guaranteing safety and reducing the algorithmic conservatism\footnote{Conceptually, it is analogues to solve an inequality-constrained optimization such as $\min_x x \ \rm{s.t} \ x \ge 0.$ }.  

\begin{figure}
    \centering
    \begin{subfigure}{0.48\textwidth}
        \centering
        \includegraphics[height=5.2cm]{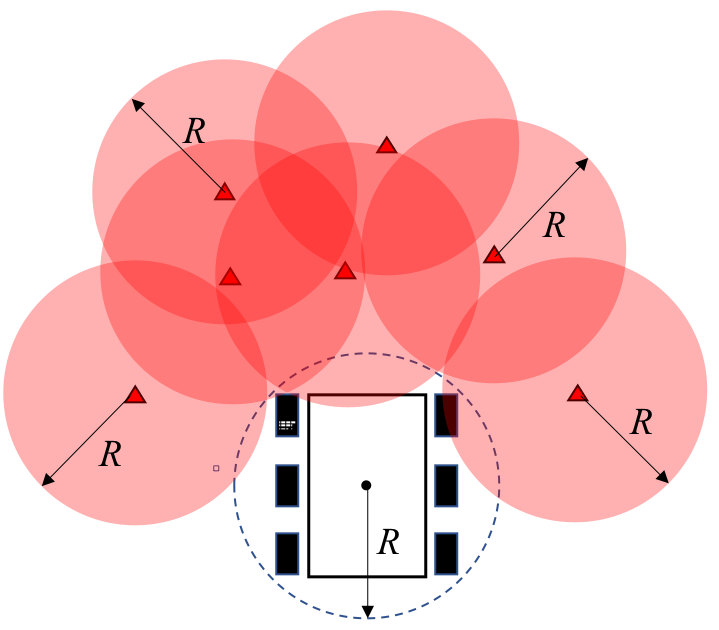}
        \caption{}
    \end{subfigure}
    \begin{subfigure}{0.48\textwidth}
        \centering
        \includegraphics[height=5.2cm]{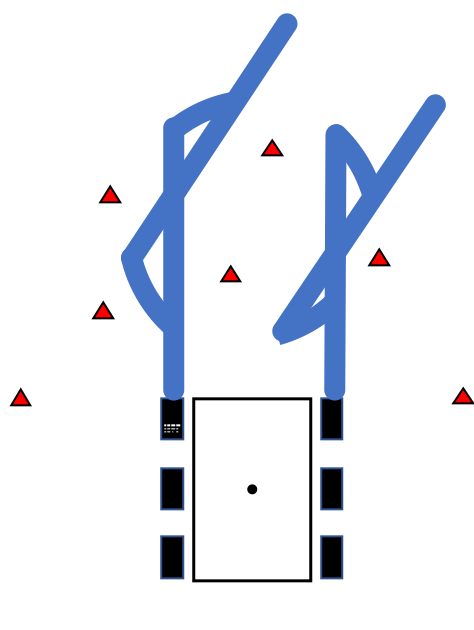}
        \caption{}
    \end{subfigure}
    \caption{Conceptual illustration of the collision checking approach in (a) GESTALT, the state-of-the-art Mars rover autonomous navigation algorithm used in Spirit, Opportunity, and Curiosity rovers, and (b) ACE, the proposed approach used in the Mars 2020 rover. The red triangles represent obstacles. GESTALT conservatively expands obstacles by the radius of the rover while ACE assesses the collision in consideration of the wheel footprints.}
    \label{fig:collision_check}
\end{figure}

We found a source of excessive conservatism in GESTALT is collision-checking. 
Most tree- or graph-based path planners, including GESTALT, need to check if each arc (edge) of the path tree or graph is safe by running a collision check at a certain interval (typically tens of cm). The collision check algorithm estimates multiple safety metrics, such as the ground clearance, tilt, and suspension angles, and check if all of them are within pre-specfied safety ranges. 
In GESTALT, the rover state is simply represented by a point in the 2D space, which represents the geometric center of the 2D footprint of the rover. 
Roughly speaking, GESTALT expands potentially colliding obstacles by the radius of the rover such that any part of the rover is guaranteed to be safe as long as the center point is outside of the expanded obstacles, as in \autoref{fig:collision_check}(a)\footnote{In reality, the terrain assessment of GESTALT is not binary; the terrain assessment map (called the \textit{goodness map}) is pre-processed by taking the worst value within the diameter of the rover from each grid of the map (i.e., dilation in image processing). This is equivalent to obstacle expansion in case of binary goodness value.}. 
Densely populated rocks may block a significant portion of the state space, resulting in a failure of finding a feasible path.
In particular, this approach does not allow the rover to straddle over a rock even if it does not hit the belly pan.
This approach is safe and computationally simple, but often overly conservative, particularly on a terrain with high rock density or undulation. 

The main idea of the proposed approach in this paper, called \textit{\ac{ACE}}, is to check collision \textit{without} expanding obstacles.
Instead of representing the rover state just by a 2D point, ACE considers the wheel footprint and resulting suspension angles to evaluate the safety metrics such as ground clearance and tilt, as illustrated in \autoref{fig:collision_check}(b). 
This approach can significantly mitigate the level of conservatism while still guaranteeing safety. 
For example, path planning with ACE allows straddling over rocks as long as there is sufficient clearance to the belly pan. 
However, precisely evaluating these metrics requires solving an iterative kinematics problem, which is not tractable given the very limited computational resources of the planetary rovers. Besides the nonlinear kinematics equations associated with the suspension mechanisms, a rough terrain profile makes it difficult to precisely predict wheel-terrain contact \cite{Sohl2005}. 
There are no known analytic solutions in general, and the problem is typically approached by iterative numerical methods at the cost of computational efficiency.
Therefore, we turn to a conservative approximation. That is, instead of running an iterative kinematic computation, ACE computes the worst-case bounds on the metrics. 
This approach is a practical compromise for our Mars rovers as it has guaranteed conservatism with acceptable computational requirement.
This claim will be empirically supported by simulations (Section \ref{sec:simulation}) and hardware experiments (Section \ref{sec:experiments}).
Furthermore, as we will empirically show in Section \ref{sec:result_conservatism}, ACE-based path planning is significantly less conservative than GESTALT.

There have been a significant body of works in literature, but none of these were sufficient for our application in terms of speed, path efficiency, and safety guarantee. Most of the motion planning methods for generic ground vehicles do not explicitly consider suspension articulation. In non-planar terrain, it is very common to model the terrain as a 2.5D or 3D map and fit a robot-sized planar patch to terrain models to obtain geometric traversability \cite{Gennery1999,Chilian2009,Ishigami2013,Wermelinger2016,Krusi2017}. It is also common to add other criteria such as roughness and step hazards to capture obstacles and undulations. Similar to GESTALT, those planners will suffer from extremely less path options in a highly cluttered environment such as the surface of Mars. Without reasonable vehicle state prediction, it is difficult to utilize the greater body-ground clearance of off-road vehicles.

To enable more aggressive yet safe planning, pose estimation on uneven terrains has been used together with path planners. Kinematics and dynamics are two major categories which account for the state of articulated models on uneven terrain. With kinematics-based approach, the problem is to find contact points between the wheel and the terrain under the kinematic constraints of the vehicle. Generic kinematics modeling is introduced for articulated rovers such as NASA/JPL's rocker-bogie rovers \cite{Tarokh2005,Chang2006}. These kinematic models are used to compute vehicle settling on uneven terrain by minimizing the wheel-ground contact errors \cite{Tarokh2005,Howard2007,Ma2019}. The terrain settling technique is used in the current ground operation of Mars rovers to check safety before sending mobility commands \cite{Yen2004,Wright2005rsvp}. The kinematic settling is also effective for other types of vehicles, such as tracked vehicles \cite{Jun2016}. The kinematics approach is generally faster than dynamics-based approach, but still computationally demanding for onboard execution on spacecraft computers.

Dynamics simulation is typically performed with general purpose physics engines. Due to its popularity, many works use \ac{ODE} for simulating robot motion during planning \cite{Papadakis2012}. \ac{ROAMS} \cite{Jain2003,Jain2004} is simulation software developed at \ac{JPL}, which models the full dynamics of flight systems including Mars rovers. \ac{ROAMS} was used to predict the high-fidelity rover behavior in rough terrain \cite{Huntsberger2008,Helmick2009}. Another faster dynamics simulator is proposed in \cite{Seegmiller2016}, which runs simulation over 1000 times faster than real time in decent computing environment. 
Although these methods can provide high-fidelity estimate in clearance, vehicle attitude, and suspension angles, they cannot directly be deployed onto the rovers due to its intractable computational cost.
Moreover, for on-board path planning in planetary missions, conservatism in safety is more important than accuracy: a single collision can terminate a mission as it is not possible to repair a damaged vehicle on another planet at least for the foreseeable future.

The main contribution of this paper is to introduce a novel kinematics solution named \ac{ACE} and its empirical validation based on simulations and hardware experiments. The key concept of \ac{ACE} is to quickly compute the vehicle configuration bounds, instead of solving the full kinematic rover-terrain settling. Knowing the bounds of certain key states, \ac{ACE} can effectively produce a conservative estimation of the rover-terrain clearance, rover attitude, and suspension angles in a closed form. \ac{ACE} is being developed as part of the autonomous surface navigation software of NASA/JPL's \ac{M2020} mission. The initial idea of \ac{ACE} appears in \cite{Otsu2016phd}, and a probabilistic extension of this work is reported in \cite{Ghosh2018}. This paper introduces improved mathematical formulation and extensive Verification and Validation (V\&V) work.

The remainder of this paper is structured as follows: \autoref{sec:model} formulates the kinematics models of articulated suspension systems, \autoref{sec:algorithm} describes the \ac{ACE} algorithm, \autoref{sec:evaluation} provides experimental results including benchmarking, and \autoref{sec:conclusion} concludes the paper.

\section{Suspension Models} \label{sec:model}

Our approach is to use kinematic equations to propagate the bounds on the height of wheels to the bounds on vehicle configuration. 
While this approach is applicable to many vehicles with articulated suspension systems used in the planetary rover domain, this section particularly focuses on the rocker and rocker-bogie suspensions. The latter is the suspension system of choice for the successful NASA/JPL's Mars rover missions \cite{Harrington2004}.

\subsection{Frames}

\begin{figure}[t]
    \centering
    \begin{subfigure}{0.48\textwidth}
        \centering
        \includegraphics[height=4.2cm]{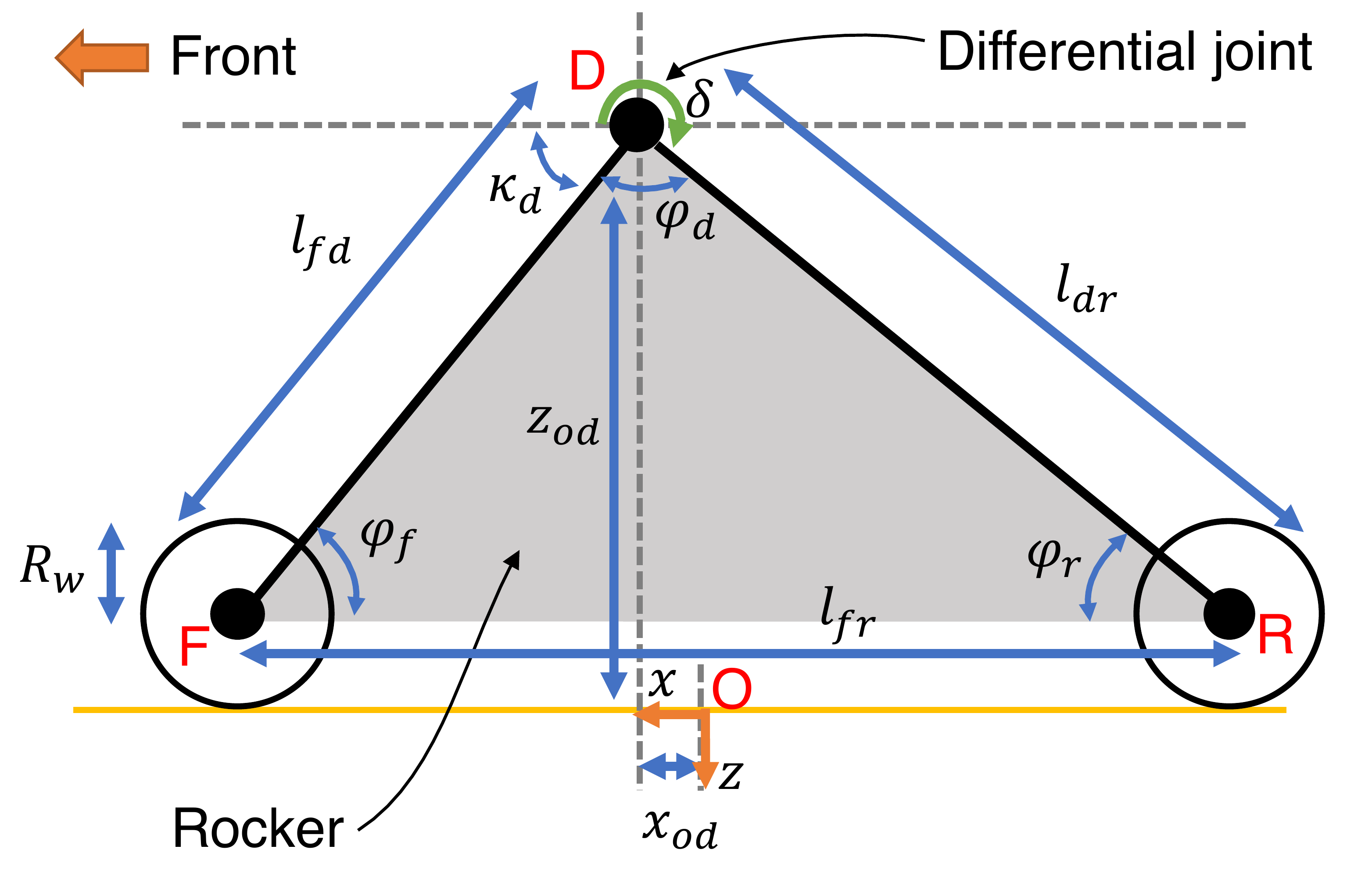}
        \caption{Rocker suspension}
        \label{fig:rocker_model}
    \end{subfigure}
    \begin{subfigure}{0.48\textwidth}
        \centering
        \includegraphics[height=4.2cm]{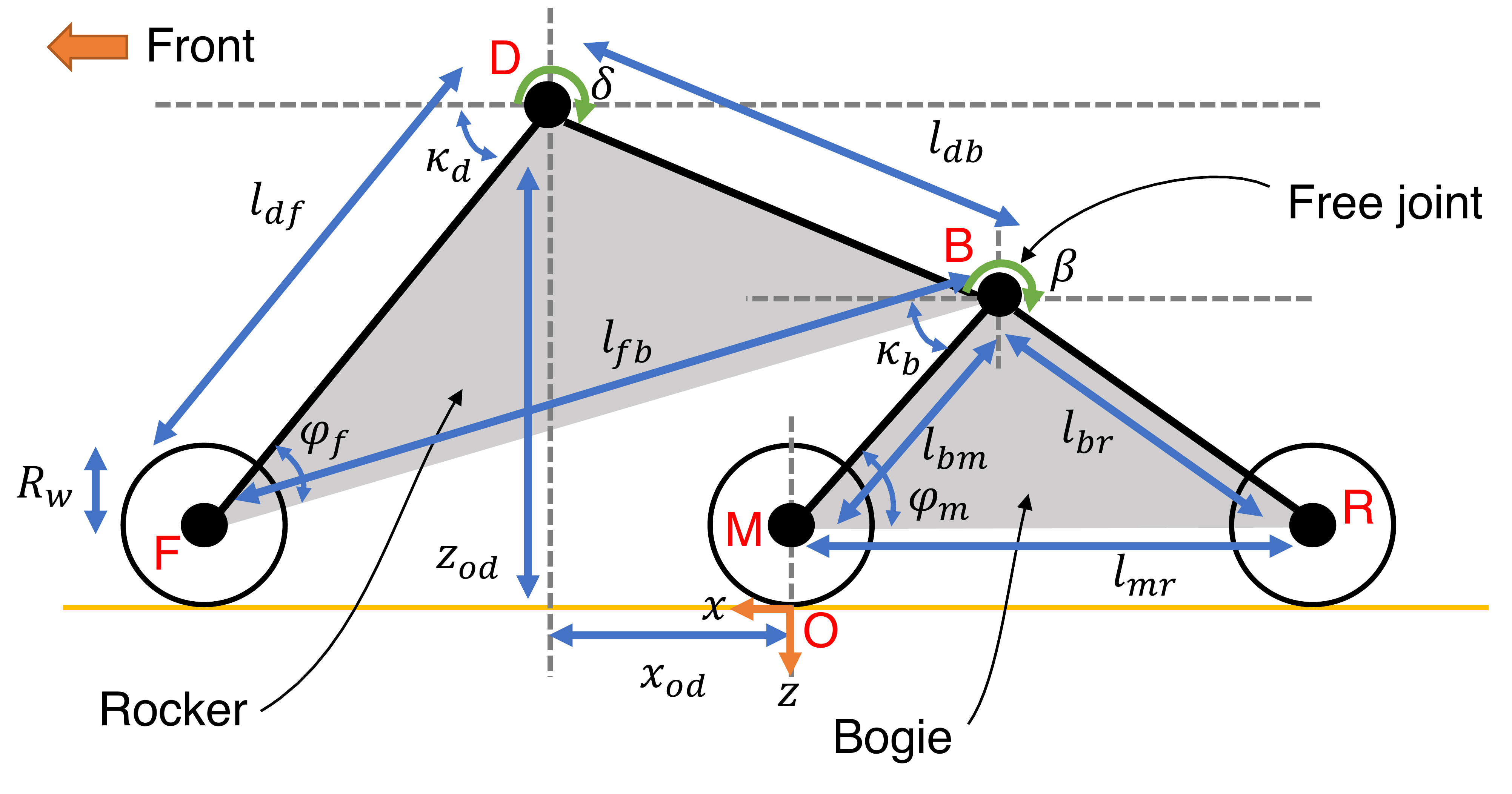}
        \caption{Rocker-bogie suspension}
        \label{fig:rockerbogie_model}
    \end{subfigure}
    \caption{Articulated suspension systems on flat ground, viewed from the left side.}
    \label{fig:suspension_models}
\end{figure}

\begin{figure}[t]
\centering
\begin{minipage}{0.45\textwidth}
    \centering
    \includegraphics[width=0.9\textwidth]{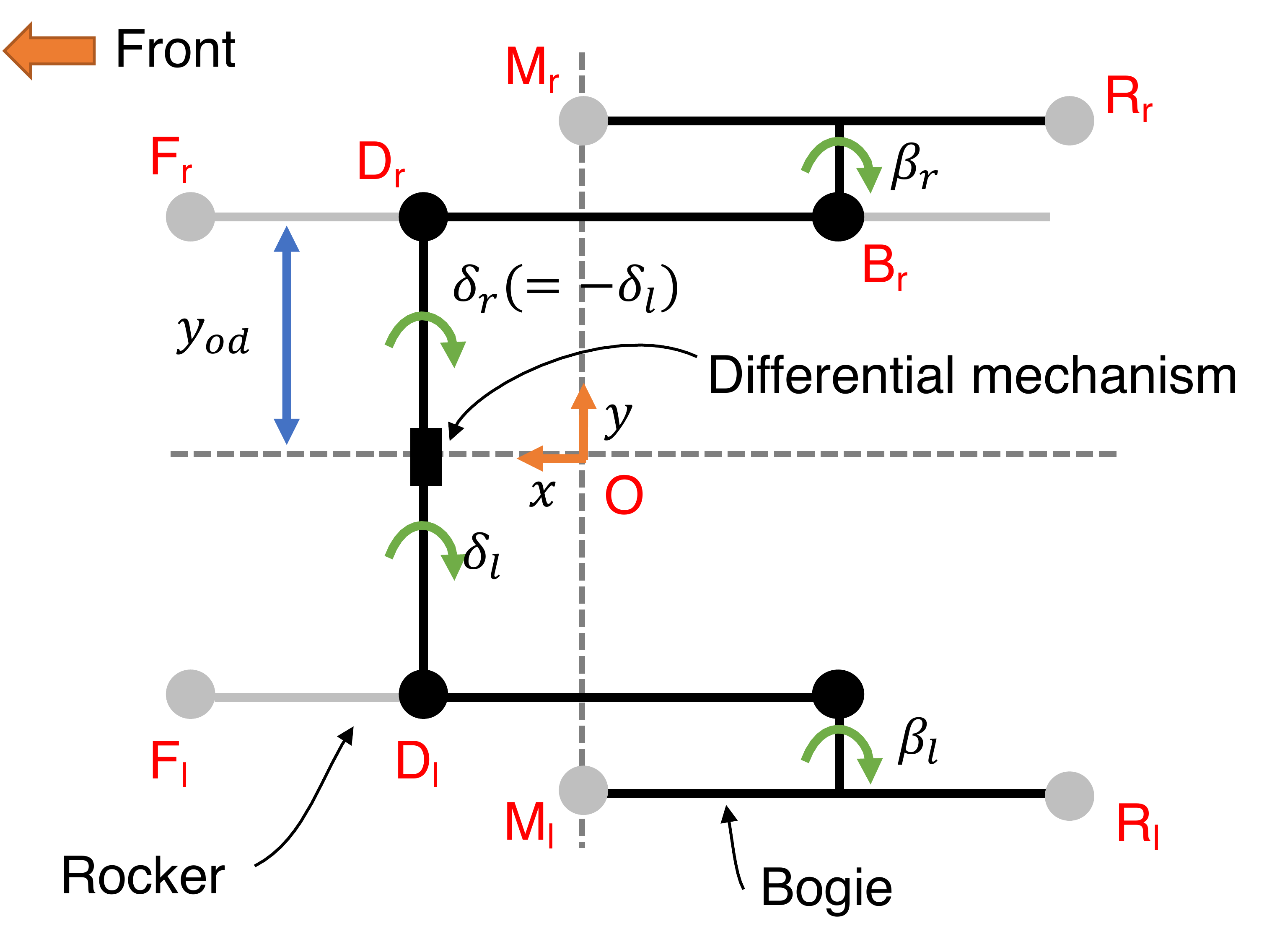}
    \caption{Rocker-bogie suspension model, viewed from top. For visibility, wheels do not represent actual positions.}
    \label{fig:rockerbogie_model_top}
\end{minipage}\hspace{5mm}
\begin{minipage}{0.45\textwidth}
    \centering
    \includegraphics[width=0.9\textwidth]{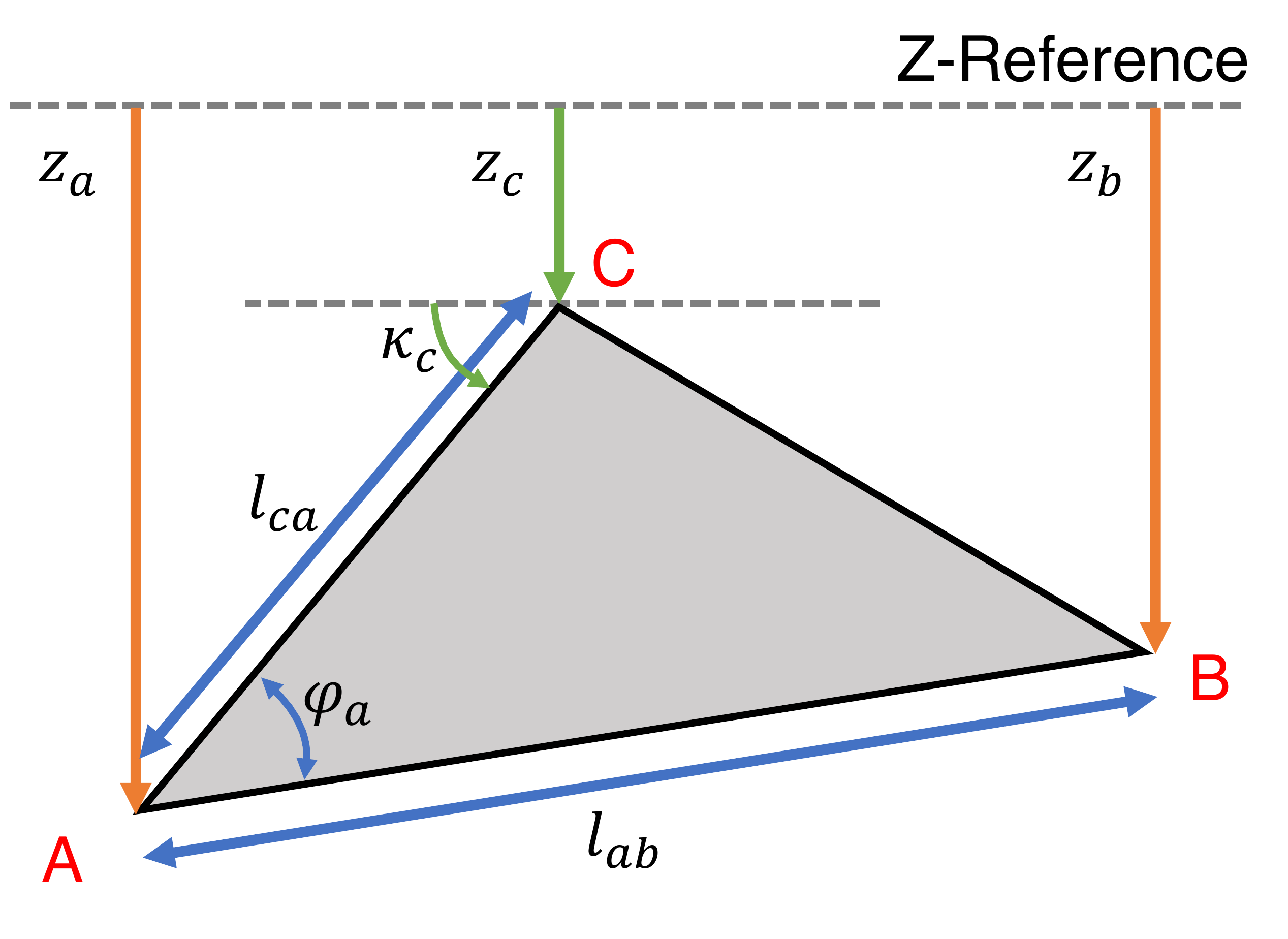}
    \caption{Simple triangular geometry that models rocker/bogie systems.}
    \label{fig:triangular_geometry}
\end{minipage}
\end{figure}

We first introduce the reference frames used in the paper, which are illustrated in \autoref{fig:suspension_models} and \ref{fig:rockerbogie_model_top}. Following the aerospace convention, the forward-right-down coordinate system is employed for the body frame of the rover. The origin is set to the center of middle wheels at the height of ground contact point when the rover is stationary on the flat ground. In this frame, the wheel heights are described in $z$-axis pointing downward (i.e., A greater wheel ``height" indicates that the wheel is moved downward).

A global reference frame is defined as a north-east-down coordinate system. The terrain geometry, which can be specified in any format such as a point cloud or a \ac{DEM}, is expressed in an arbitrary frame. 
The rover path planning is conventionally performed in 2D or 2.5D space based on the nature of rover's mobility systems. A path is typically represented as a collection of poses containing 2D position and heading angle $ (x, y, \psi) $. A path is regarded as safe if all poses along the path satisfies the safety constraints.

\subsection{Rocker Suspension}

The rocker suspension in \autoref{fig:rocker_model} is a simpler variant of the rocker-bogie suspension, which will be discussed in the next section. The rocker suspension usually consists of four wheels, where the two wheels on the same side are connected with a rigid rocker link. The left and right suspensions are related through a passive differential mechanism, which transfers a positive angle change on one side as a negative change to the other side.

The kinematic relation of the rocker suspension is represented by a simple triangular geometry in \autoref{fig:triangular_geometry}. Consider a triangle ABC with a known shape parameterized by two side lengths and the angle between them $ (l_{ca}, l_{ab}, \varphi_{a}) $. Given the height of A and B (i.e., $ z_{a}, z_{b} $), there are up to four solutions for $ z_{c} $, but other constraints such as vehicle orientation uniquely specifies a single solution given by:
\begin{align}
    z_{c} = z_{a} - l_{ca}\sin \kappa(z_{a}, z_{b})
    \label{eq:tri_z}
\end{align}
where $ \kappa(\cdot) $ denotes an angle of link AC with respect to the reference line (i.e., $\kappa_c$) and is defined as
\begin{align}
    \kappa(z_a, z_b) = 
    \varphi_a + \asin\left(\cfrac{z_{a}-z_{b}}{l_{ab}}\right)
    \:.
    \label{eq:tri_kappa}
\end{align}
The solution only exists if $ | z_{a}-z_{b} | \leq l_{ab} $.

The rocker suspension model is formulated using the triangular geometry in \eqref{eq:tri_z} and \eqref{eq:tri_kappa}. Given two wheel heights $ z_{f} $ and $ z_{r} $, the rocker joint height is given as
\begin{align}
    z_{d} = z_{f} - l_{df}\sin \kappa_d(z_{f}, z_{r})
\end{align}
where $ l_{df} $ is the length of front rocker link and $ \kappa_d(\cdot) $ is an instance of \eqref{eq:tri_kappa} with the rocker suspension parameters $ (l_{df}, l_{fr}, \varphi_{f}) $. Due to the differential mechanism, the left and right rocker angles in relative to the body, $\delta_{l}, \delta_{r}$, have the same absolute value with the opposite sign. They can be computed from link angles as:
\begin{align}
    \delta_{l} = -\delta_{r} 
    = \cfrac{\kappa_d(z_{f_r}, z_{r_r}) - \kappa_d(z_{f_l}, z_{r_l}) }{2}
    \:.
\end{align}

The body attitude is a function of left and right rocker joint states. The roll angle of the body is computed from the difference of joint heights:
\begin{align}
    \phi &= \asin\left(
    \cfrac{z_{d_r} - z_{d_l}}{2 y_{od}}
    \right),
\end{align}
where $ y_{od} $ is the lateral offset from the center of body to a differential joint. The pitch angle is computed as 
\begin{align}
    \theta &= \kappa_{d0} - \cfrac{\kappa_d(z_{f_l}, z_{r_l})+\kappa_d(z_{f_r}, z_{r_r})}{2}
\end{align}
where the first term represents an angle offset of front link when the rover is on a flat ground ($ \kappa_{d0}=\varphi_f $ in this example).

Finally, the body frame height in the global frame can be obtained as
\begin{align}
    z_o = \cfrac{z_{d_{l}} + z_{d_{r}}}{2}
    + x_{od} \sin\theta\cos\phi - z_{od} \cos\theta \cos\phi
\end{align}
where $ x_{od} $ and $ z_{od} $ are offsets from the body frame origin to a differential joint. Since the belly pan is rigidly attached to the body frame, the rover-terrain clearance can be derived from these height and attitude information.

\subsection{Rocker-bogie Suspension}

The rocker-bogie suspension (\autoref{fig:rockerbogie_model}) is a rocker suspension with an additional free joint on each side. According to the previous Mars rover conventions, we assume that the front wheels of the six-wheeled rover are connected directly to the rocker suspension while the middle and rear wheels are attached to the bogie suspension. The inverse kinematics of the rocker-bogie suspension can be derived by extending that of the rocker suspension, described in the previous subsection.

We first determine the state of bogie link. The bogie joint heights can be estimated from middle and rear wheel heights $ (z_{m}, z_{r}) $
\begin{align}
    z_{b} = z_{m} - l_{bm}\sin \kappa_b(z_{m}, z_{r})
\end{align}
where $l_{bm}$ is the length of bogie front link and $ \kappa_b(\cdot) $ denotes the triangular geometry for the bogie triangle. Using the height of bogie joint $ z_{b} $, the rocker joint height can be computed as 
\begin{align}
    z_{d} = z_{f} - l_{df}\sin \kappa_d(z_{f}, z_{b})
    \:.
\end{align}

Given the heights of the wheels and joints, rocker and bogie angle changes are computed as
\begin{align}
    \delta_{l} &= -\delta_{r} 
    = \cfrac{\kappa_d(z_{f_{r}}, z_{b_{r}}) - \kappa_d(z_{f_{l}}, z_{b_{l}})}{2} \\
    \beta_{l} &= 
    \kappa_d(z_{f_{l}}, z_{b_{l}}) -
    \kappa_b(z_{m_{l}}, z_{r_{l}}) - \kappa_{d0} + \kappa_{b0} \\ 
    \beta_{r} &= 
    \kappa_d(z_{f_{r}}, z_{b_{r}}) -
    \kappa_b(z_{m_{r}}, z_{r_{r}}) - \kappa_{d0} + \kappa_{b0}
\end{align}
where $ \kappa_{d0} $ and $ \kappa_{b0} $ denote the initial angles of rocker and bogie joints. 
The attitude and height of the body are derived as:
\begin{align}
    \phi &= \asin\left(
    \cfrac{z_{d_r} - z_{d_l}}{2y_{od}}
    \right) \label{eq:roll} \\
    \theta &= \kappa_{d0} - \cfrac{\kappa_d(z_{f_l}, z_{b_l})+\kappa_d(z_{f_r}, z_{b_r})}{2} \label{eq:pitch} \\
    z_o &= \cfrac{z_{d_{l}} + z_{d_{r}}}{2}
    + x_{od} \sin\theta \cos\phi - z_{od} \cos\theta \cos\phi
    \label{eq:z_o}
    \:.
\end{align}

\section{Algorithm} \label{sec:algorithm}

\begin{figure}
    \centering
    \includegraphics[width=0.5\textwidth]{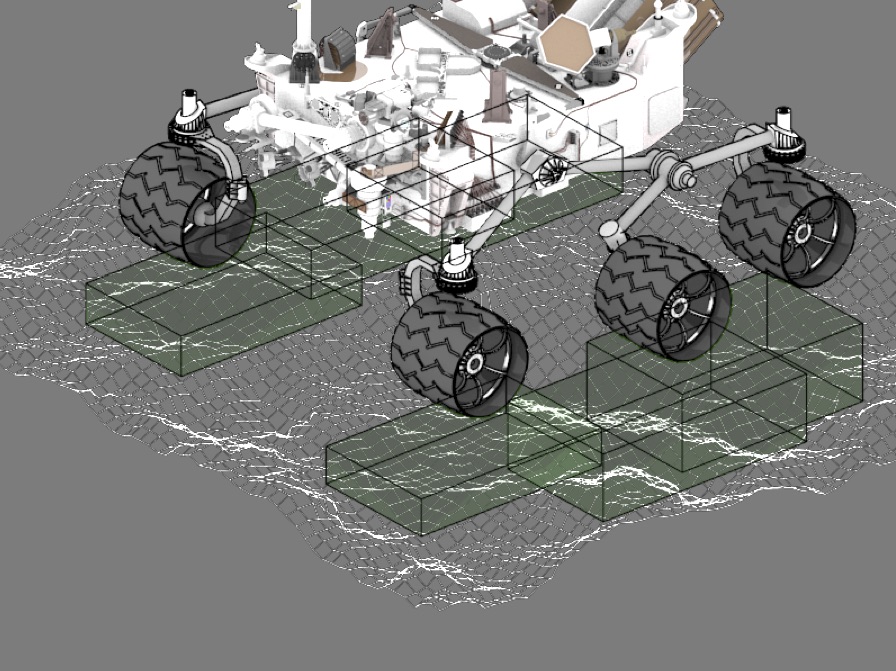}
    \caption{Ranges of possible wheel configurations computed from terrain geometry and mechanical constraints.}
    \label{fig:wheel_intervals}
\end{figure}

Remember that \ac{ACE} is designed to quickly compute conservative bounds on vehicle states. 
Unlike the full kinematics settling that relies on iterative numerical methods, 
our approach computes the bounds in a closed form. The ACE algorithm is summarized as follows:
\begin{enumerate}
    \item　For a given target rover pose $ (x, y, \psi) $, find a rectangular wheel box in x-y plane in the body frame that conservatively includes the footprint of each wheel over any possible rover attitude and suspension angles. 
    \item Find the minimum and maximum terrain heights in each of the wheel boxes (see \autoref{fig:wheel_intervals}).
    \item Propagate the bounds on wheel heights to the vehicle configuration with the kinematic formula derived in the previous section.
    \item Assess vehicle safety based on the worst-case states.
\end{enumerate}
In 3), all possible combinations are considered to obtain the worst-case bounds. Due to the monotonic nature of suspension, the bounds can be obtained via the evaluation of extreme configurations. For example, the bounds on the rocker/bogie states are obtained by finding the worst cases among the eight extreme combinations of the min/max heights of three wheels, as illustrated in \autoref{fig:extreme_cases}. This propagation process is visually presented in the supplemental video. 

To precisely describe the algorithm, we first introduce the interval arithmetic as a mathematical framework in our method. We then describe how we apply it to solve our problem with a case study using the \ac{M2020} rover.

\subsection{Notation}

In the interval arithmetic~\cite{Hickey2001}, an \textit{interval} is defined as follows
\begin{align}
    \interval{x} = \{ x\in\mathbb{R}^{*} \mid x^{-} \leq x \leq x^{+} \}
\end{align}
where a pair $ \interval{x} $ represents the all reals between two. The symbol $ \mathbb{R}^{*} $ denotes an extended real defined as $ \mathbb{R}^{*} = \mathbb{R} {\cup} \{-\infty, \infty\} $.
Elementary arithmetic operations on reals can be extended to intervals, such as
\begin{align}
    \interval{x} + \interval{y} &= [x^{-}+y^{-},\ x^{+}+y^{+}] \\
    \interval{x} - \interval{y} &= [x^{-}-y^{+},\ x^{+}-y^{-}] .
\end{align}

For a continuous function $ f(x) $, we can extend its input and output space to intervals
\begin{align}
    f(\interval{x}) = \left[
    \min_{x\in\interval{x}}f(x),\ \max_{x\in\interval{x}}f(x)
    \right]
    \:.
\end{align}
Computing the minimum and maximum is trivial if the function $ f $ is monotonic, or special non-monotonic functions such as trigonometric functions.

In the rest of the paper, we use the following abbreviation to represent an interval unless explicitly stated
\begin{align}
    [x] \equiv \interval{x}  
    \:.
\end{align}

\subsection{Wheel Height Intervals}

\begin{figure}
    \centering
    \begin{subfigure}{0.38\textwidth}
        \centering
        \includegraphics[width=\textwidth]{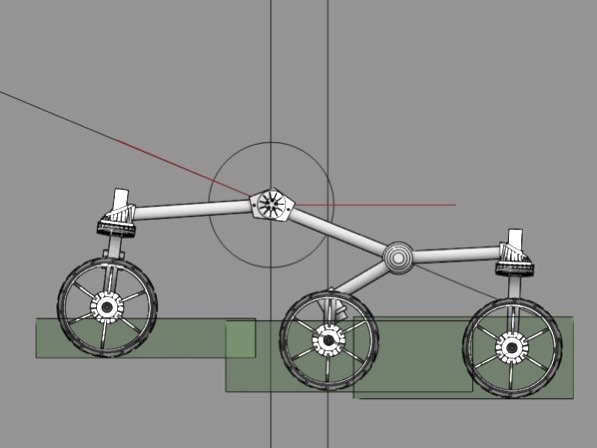}
        \caption{$(z_{f}^{+}, z_{m}^{+}, z_{r}^{+})$}
    \end{subfigure}
    \begin{subfigure}{0.38\textwidth}
        \centering
        \includegraphics[width=\textwidth]{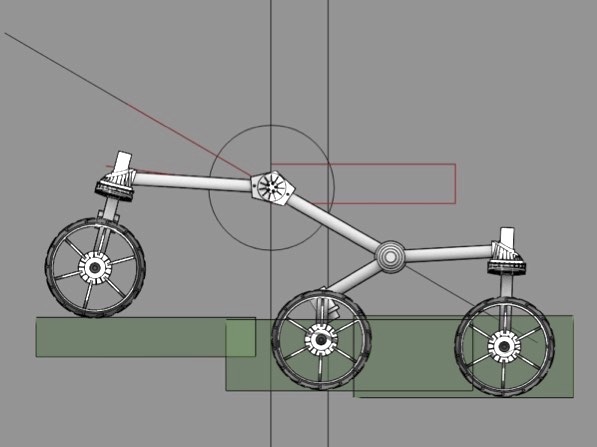}
        \caption{$(z_{f}^{-}, z_{m}^{+}, z_{r}^{+})$}
    \end{subfigure}
    \begin{subfigure}{0.38\textwidth}
        \centering
        \includegraphics[width=\textwidth]{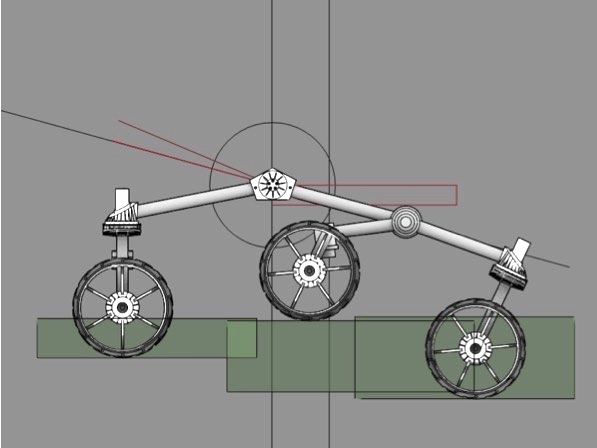}
        \caption{$(z_{f}^{+}, z_{m}^{-}, z_{r}^{+})$}
    \end{subfigure}
    \begin{subfigure}{0.38\textwidth}
        \centering
        \includegraphics[width=\textwidth]{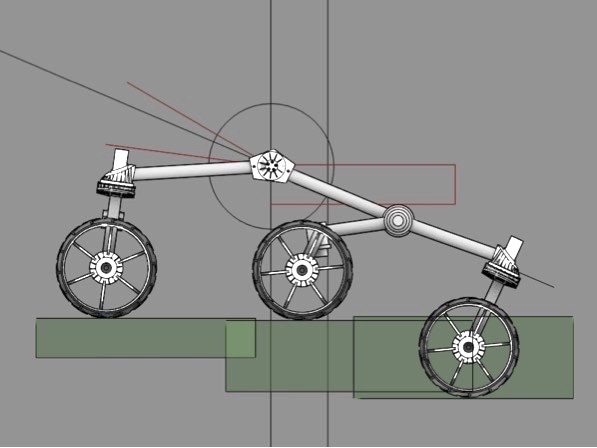}
        \caption{$(z_{f}^{-}, z_{m}^{-}, z_{r}^{+})$}
    \end{subfigure}
    \begin{subfigure}{0.38\textwidth}
        \centering
        \includegraphics[width=\textwidth]{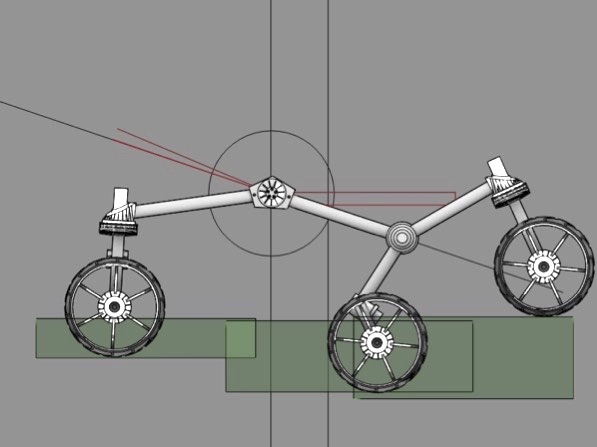}
        \caption{$(z_{f}^{+}, z_{m}^{+}, z_{r}^{-})$}
    \end{subfigure}
    \begin{subfigure}{0.38\textwidth}
        \centering
        \includegraphics[width=\textwidth]{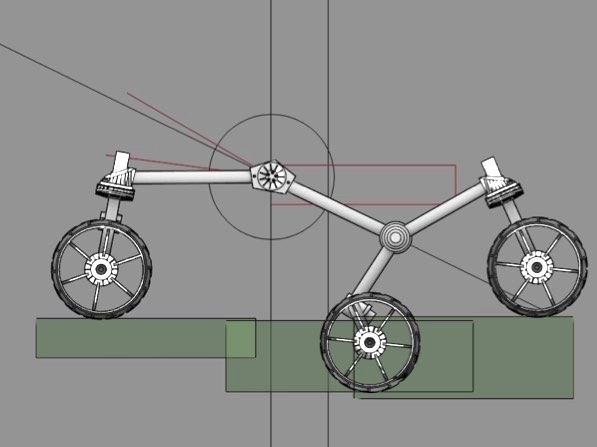}
        \caption{$(z_{f}^{-}, z_{m}^{+}, z_{r}^{-})$}
    \end{subfigure}
    \begin{subfigure}{0.38\textwidth}
        \centering
        \includegraphics[width=\textwidth]{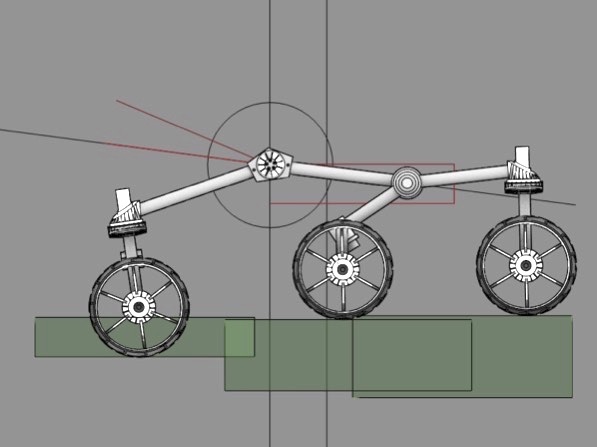}
        \caption{$(z_{f}^{+}, z_{m}^{-}, z_{r}^{-})$}
    \end{subfigure}
    \begin{subfigure}{0.38\textwidth}
        \centering
        \includegraphics[width=\textwidth]{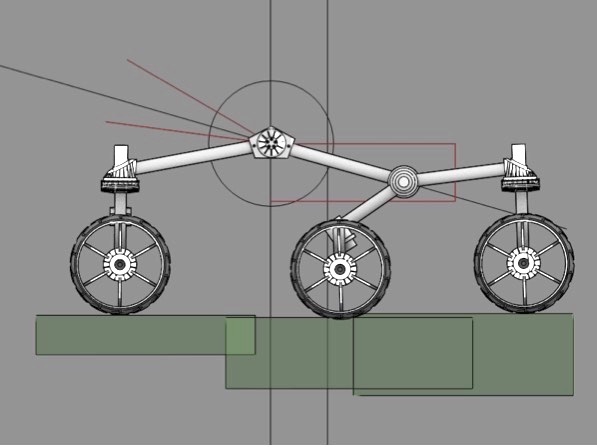}
        \caption{$(z_{f}^{-}, z_{m}^{-}, z_{r}^{-})$}
    \end{subfigure}
    \caption{Extreme configurations for state bound computation. Three elements in a tuple represent the front, middle and rear wheel heights, respectively. The superscripts $+$ and $-$ represent the maximum and minimum for the state variable.}
    \label{fig:extreme_cases}  
\end{figure}

Firstly, we estimate the wheel height intervals based on terrain measurements from sensors (e.g., stereo vision). The span of wheel heights can be computed from the highest and lowest terrain points within the wheel boxes (see green boxes in \autoref{fig:wheel_intervals}). The $x$ and $y$ dimensions of the wheel boxes are derived from the vehicle's mechanical properties such as wheel size, suspension constraints, and vehicle tip-over stability. We can determine a conservative range of wheel contact locations for all possible suspension angles and stable attitude.

In the rest of this paper, we represent the bound for $i$-th wheel by $ [z_i] $. It is important to estimate these bounds conservatively to make the final state bounds to be complete, since the uncertainty in wheel heights is directly propagated to other states. For the conservative estimate, we may need to include the dynamic effect such as terrain deformation, wheel slips and sinkage, depending on the environment to explore. In addition, it is important to consider the effect of perception error as detailed in the experiment section.

\subsection{Suspension Intervals}

Since $\asin(x)$ is monotonically increasing in $ x \in [-1, 1] $, we can extend the concept of intervals to the function $ \kappa(\cdot) $ in \eqref{eq:tri_kappa}
\begin{align}\label{eq:21}
    \kappa([z_{a}], [z_{b}]) = [
    \kappa(z_{a}^{-}, z_{b}^{+}), \kappa(z_{a}^{+}, z_{b}^{-})
    ] .
\end{align}
On the other hand, \eqref{eq:tri_z} is a convex function which has a global minimum if $ z_{b} = z_{b}^{-} $ and the rear link CB is aligned with $z$-axis. In case of the \ac{M2020} rover, the minimum is located outside of the mechanical limits. Therefore, in practice, we can assume the monotonicity and use the following interval for the height
\begin{align}
    [z_{c}] &= [
    z_{a}^{-} - l_{ab}\sin \kappa_b(z_{a}^{-}, z_{b}^{-}),\ 
    z_{a}^{+} - l_{ab}\sin \kappa_b(z_{a}^{+}, z_{b}^{+})
    ]
    \:.
\end{align}

Based on the above intervals, the suspension state intervals are computed for joint heights
\begin{align}
    [z_{b}] &= [
    z_{m}^{-} - l_{bm}\sin \kappa_b(z_{m}^{-}, z_{r}^{-}),\ 
    z_{m}^{+} - l_{bm}\sin \kappa_b(z_{m}^{+}, z_{r}^{+})
    ] \\
    [z_{d}] &= [
    z_{f}^{-} - l_{df}\sin \kappa_d(z_{f}^{-}, z_{b}^{-}),\ 
    z_{f}^{+} - l_{df}\sin \kappa_d(z_{f}^{+}, z_{b}^{+})
    ]
\end{align}
and for joint angles
\begin{align}
    [\delta_l] &= -[\delta_r] 
    = \left[
    \cfrac{\kappa_d(z_{f_{r}}^{-}, z_{b_{r}}^{+}) - \kappa_d(z_{f_{l}}^{+}, z_{b_{l}}^{-})}{2},\ 
    \cfrac{\kappa_d(z_{f_{r}}^{+}, z_{b_{r}}^{-}) - \kappa_d(z_{f_{l}}^{-}, z_{b_{l}}^{+})}{2}
    \right] \\
    [\beta_{l}] &\subseteq \left[
        \kappa_d(z_{f_{l}}^{-}, z_{b_{l}}^{+}) -
        \kappa_b(z_{m_{l}}^{+}, z_{r_{l}}^{-}) - \kappa_{d0} + \kappa_{b0},\ 
        \kappa_d(z_{f_{l}}^{+}, z_{b_{l}}^{-}) -
        \kappa_b(z_{m_{l}}^{-}, z_{r_{l}}^{+}) - \kappa_{d0} + \kappa_{b0}
    \right] \\
    [\beta_{r}] &\subseteq \left[
        \kappa_d(z_{f_{r}}^{-}, z_{b_{r}}^{+}) -
        \kappa_b(z_{m_{r}}^{+}, z_{r_{r}}^{-}) - \kappa_{d0} + \kappa_{b0},\ 
        \kappa_d(z_{f_{r}}^{+}, z_{b_{r}}^{-}) -
        \kappa_b(z_{m_{r}}^{-}, z_{r_{r}}^{+}) - \kappa_{d0} + \kappa_{b0}
    \right] .
\end{align}
For the sake of simplicity, we use loose bounds for the bogie angles., The boundary configurations may be impossible in reality. In this example, the lower bound of $\beta$ requires $(z_{f}^{-}, z_{m}^{+}, z_{r}^{-}, z_{b}^{+})$ but $z_{b}^{+}$ requires $(z_{m}^{+}, z_{r}^{+})$, which is inconsistent in $z_{r}$ (except the case where $z_{r}^{-}$ and $z_{r}^{+}$ are identical).

\subsection{Attitude Intervals}

Similarly, the intervals for body attitude can be derived from wheel height intervals. Using the kinematics equations \eqref{eq:roll} and \eqref{eq:pitch} yields
\begin{align}
    [\phi] &= \left[
    \asin\left(\cfrac{z_{d_r}^{-} - z_{d_l}^{+}}{2y_{od}} \right),\ 
    \asin\left(\cfrac{z_{d_r}^{+} - z_{d_l}^{-}}{2y_{od}} \right)
    \right] \\
    [\theta] &= \left[
        \kappa_{d0} - \cfrac{\kappa_d(z_{f_l}^{+}, z_{b_l}^{-}) + \kappa_d(z_{f_r}^{+}, z_{b_r}^{-})}{2},\ 
        \kappa_{d0} - \cfrac{\kappa_d(z_{f_l}^{-}, z_{b_l}^{+}) + \kappa_d(z_{f_r}^{-}, z_{b_r}^{+})}{2}
    \right] .
\end{align}

\subsection{Clearance Intervals}

Since the vehicle body has connection to the world only through its suspension and wheel systems, its configuration is fully determined by the suspension state. The body height bound in the world frame is computed with \eqref{eq:z_o}:
\begin{align}
    [z_o] \subseteq &\left[
    \cfrac{z_{d_l}^{-} + z_{d_r}^{-}}{2}
    - z_{od} \cos|\theta|^{+} \cos|\phi|^{+}
    + x_{od} \min(\sin\theta^{-} \cos|\phi|^{-},\ \sin\theta^{-} \cos|\phi|^{+})
    , \right. \nonumber\\
    & \ \ 
    \left.
    \cfrac{z_{d_l}^{+} + z_{d_r}^{+}}{2} 
    - z_{od}  \cos|\theta|^{-} \cos|\phi|^{-} 
    + x_{od} \max(\sin\theta^{+} \cos|\phi|^{-},\ \sin\theta^{+} \cos|\phi|^{+})
    \right]
\end{align}
using the intervals of absolute roll/pitch angles $ [|\phi|], [|\theta|] $. Note that the trigonometric functions in the equations are monotonic since we can assume $ |\phi|,|\theta| \in \left[ 0, \tfrac{\pi}{2} \right] $ for typical rovers.
Assuming the belly pan is represented as a plane with width $w_{p}$ and length $l_{p}$ at nominal ground clearance $ c_0 $, a loose bound for the maximum (lowest) height point in belly pan is computed as
\begin{align}
    [z_{p}] \subseteq [ 
    & z_o^{-} - c_{0} \cos|\theta|^{-} \cos|\phi|^{-} 
    + \cfrac{l_{p}}{2} \sin|\theta|^{-} \cos|\phi|^{+} 
    + \cfrac{w_{p}}{2} \sin|\phi|^{-}, \nonumber\\
    & z_o^{+} - c_{0} \cos|\theta|^{+} \cos|\phi|^{+} 
    + \cfrac{l_{p}}{2} \sin|\theta|^{+} \cos|\phi|^{-} 
    + \cfrac{w_{p}}{2} \sin|\phi|^{+}
    ]
\end{align}

Let's define the rover-ground clearance as a height gap between the lowest point of the belly pan and the highest point of the ground. This is a conservative definition of clearance. Given the intervals of ground point height under the belly pan $ [z_{m}] $, the clearance is computed as
\begin{align}\label{eq:32}
    [c] &\equiv [z_{m}^{-} - z_{p}^{+},\ z_{m}^{-} - z_{p}^{-}]
    \:.
\end{align}

\subsection{Safety Metrics}

We use the above state intervals to test if a given pose has chance to violate safety conditions. Different safety conditions can be applied to different rovers. For example, the following metrics are considered for the \ac{M2020} rover.
\begin{itemize}
    \item \textit{Ground clearance} must be greater than the threshold.
    \item \textit{Body tilt} (computed from roll and pitch angles) must be smaller than the threshold.
    \item \textit{Suspension angles} must stay within the predefined safety ranges.
    \item \textit{Wheel drop} (defined as a span of wheel height uncertainty) must be smaller than the threshold.
\end{itemize}

\section{Experiments} \label{sec:evaluation}

Recall that ACE is designed to be conservative, and at the same time, to reduce the conservatism compared to the state-of-the-art. In Sections \ref{sec:simulation} and \ref{sec:experiments}, we will show that ACE is conservative, hence safe, through simulation and hardware experiments. 
Then, Section \ref{sec:result_conservatism} compares the algorithmic conservatism with the state-of-the-art. 
Our tests involve rovers with different sizes to observe the performance difference due to mechanical system configurations.

\subsection{Simulation Study}
\label{sec:simulation}

We first tested ACE with simulation to validate the algorithm in noise-free scenarios. On these tests, terrain topological models are directly loaded from the simulator to ACE. Therefore, the presented results are not contaminated by measurement noise from perception systems. The algorithm is tested on different terrain configurations, from artificial quadratic functions to a realistic Martian terrain model. We use simulator-reported rover state as ground truth, which is originally computed with iterative numeric methods based on rover and terrain models.

\subsubsection{Simulation with a single ACE run}

\begin{figure}
    \centering
    \begin{subfigure}{0.48\textwidth}
        \includegraphics[width=0.9\textwidth]{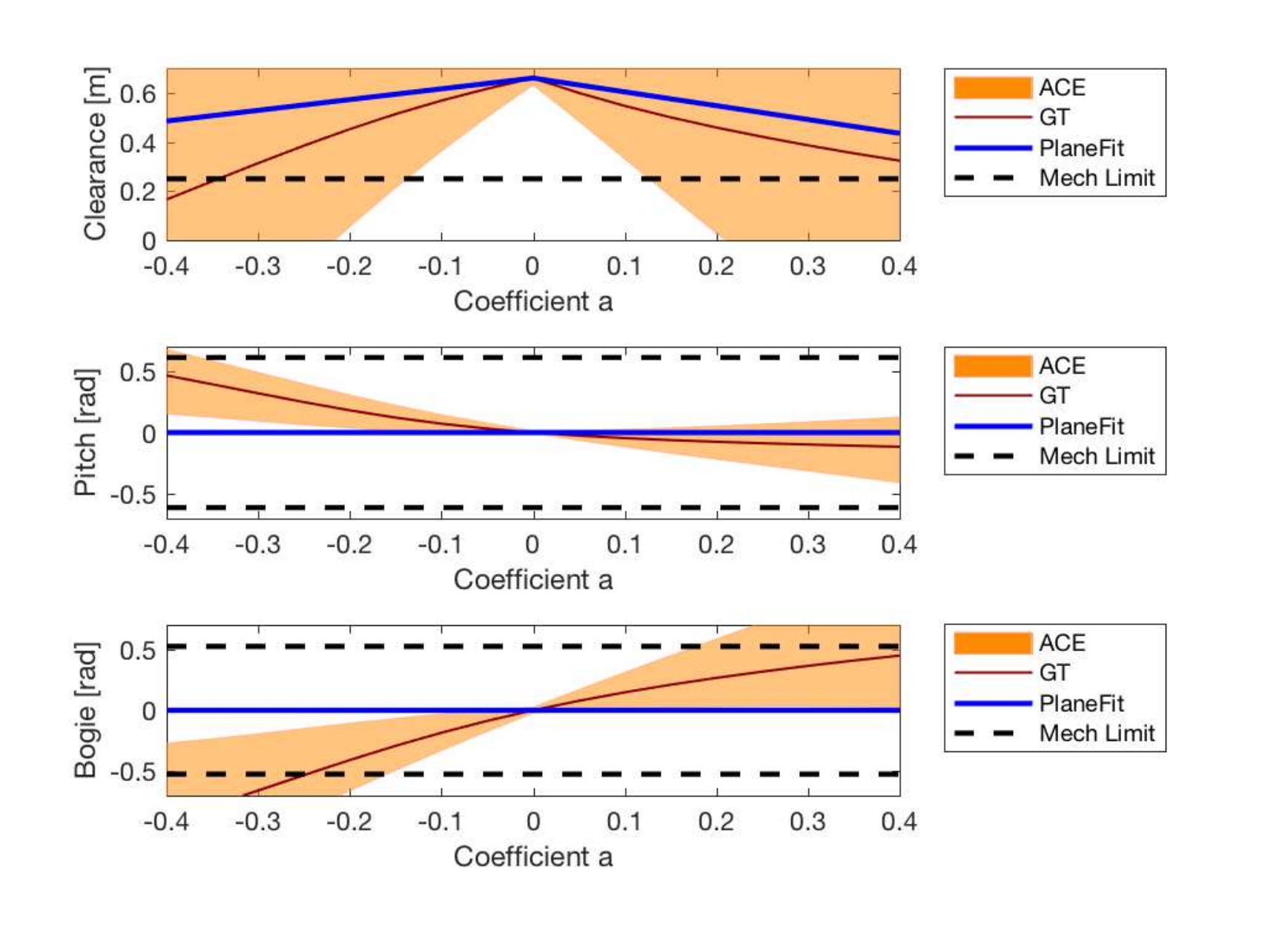}
        \caption{Ground-truth and predicted states}
    \end{subfigure}
    \begin{subfigure}{0.48\textwidth}
        \includegraphics[width=0.9\textwidth]{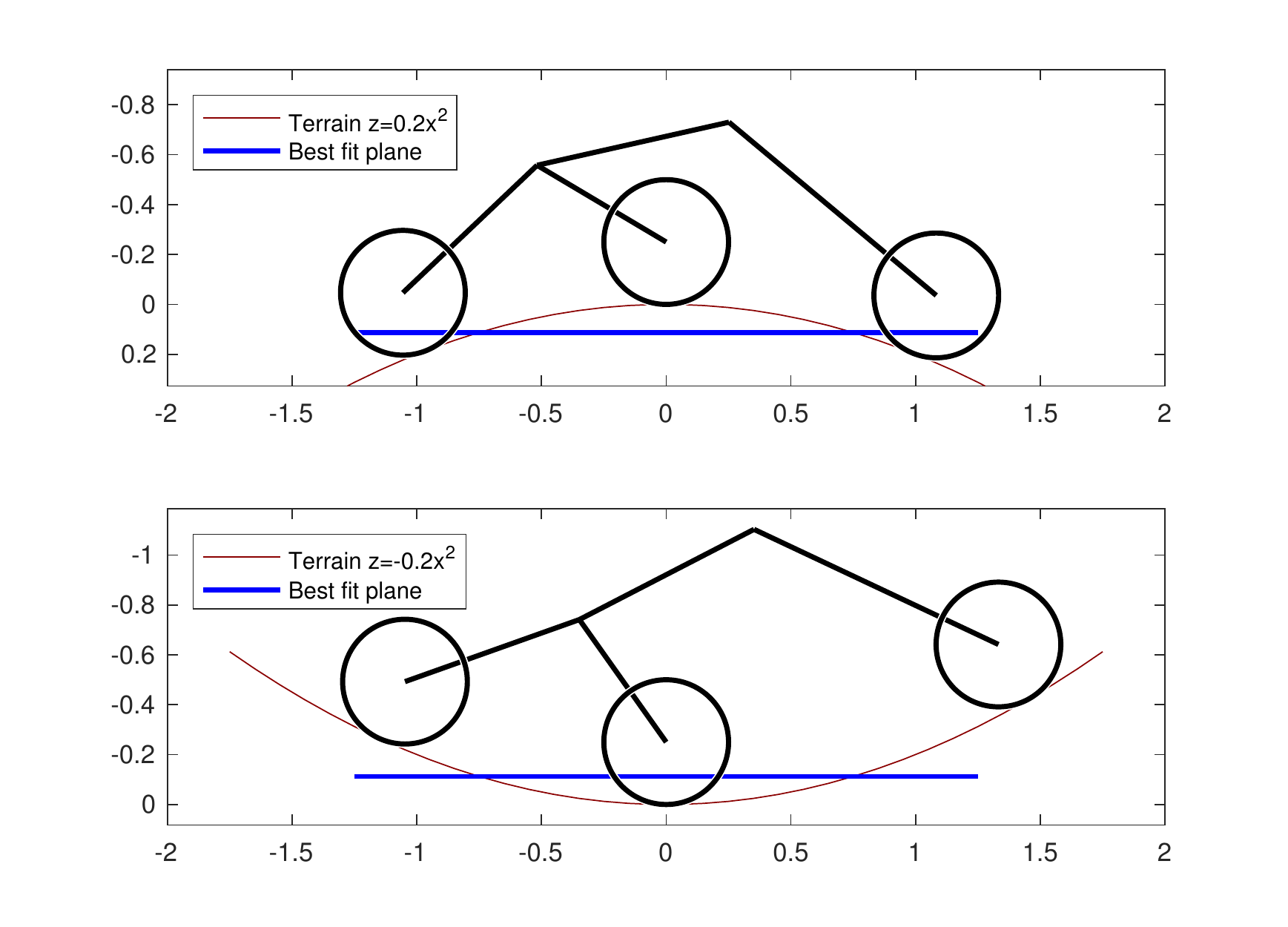}
        \caption{Side view (concave and convex)}
    \end{subfigure}
    \caption{Analysis on artificial terrains with varying undulation levels}
    \label{fig:state_undulation}
\end{figure}

We placed a simulated rover on simple geometric terrains and run ACE to compute bounds on the belly pan clearance, the attitude, and the suspension angles. 
Their ground-truth values were also obtained from the simulation to check if they are conservatively confined by the ACE bounds. 
In addition, these results were compared against a simple, planefit-based estimation approach.
More specifically, we fitted a plane to a given topography over a window with a 1.25 m radius and approximately estimate the rover's pose by assuming that the rover is placed on this plane. 
The ground clearance was estimated by computing the difference between the highest point of the terrain in the window and the belly pan height based on the estimated rover pose. 
The planefit-based approximation gives the exact ground clearance when the terrain is flat. 
We chose plane fitting as the point of comparison because, as we shall see shortly, it provides an insight to the cause of the conservatism of GESTALT, the state-of-the-art autonomous rover navigation algorithm used for the three existing Mars rovers. 

We used simple terrains represented by $z=ax^2$ in the body frame with varying $a$ for this test. 
Tests with more complex, realistic terrains will follow.
The ground-truth settling was obtained via a numeric optimization method. 

\autoref{fig:state_undulation} shows the results.
Note that the $z$-axis points downwards, meaning that the terrain is convex with a negative $a$ and concave with a positive $a$.
The brown solid lines represent ground-truth states, with ACE bounds denoted by orange shaded areas. 
As expected, ACE bounds always provided conservative estimate. 
Compared to attitude and suspension angles, the clearance estimation resulted in a greater conservatism in general. This is because the clearance is the last estimated property propagated from terrain heights and hence accumulates uncertainty. 

In contrast, the planefit-based approach consistently gave an optimistic estimate of the ground clearance and the pitch angle. 
In addition, since the rover is always placed on a plane, the bogie angle is always estimated to be zero.
GESTALT does not explicitly computes the ground clearance; instead, it computes the ``goodness" of each cell on the terrain from multiple factors including roughness (i.e., residual from the planefit) and step obstacles (i.e., maximum height difference between adjacent cells), where the weights on each factor are manually tuned such that the conservatism is guaranteed for the worst cases.
The fact that the planefit-based clearance estimation is optimistic for non-zero $a$ implies that the weights on roughness and the step obstacle must be sufficiently great for the worst-case $a$.
This in turn makes the algorithm overly conservative when the terrain is nearly flat (i.e., smaller $a$), which is the case for most of the time of driving on Mars. 
In contrast, ACE gives tighter bounds for a smaller $a$.
This illustrates a desirable behavior of ACE; that is, it adjusts the level of conservatism depending on the terrain undulation. 
It results in the exact estimation on a perfectly flat terrain and increases conservatism for undulating terrains. 
Overall, the ground truths are always conservatively bounded. 
We also note that ACE becomes overly conservative on a highly undulating terrain.
We believe the impact of this issue in practical Mars rover operation is relatively limited because we typically avoid such terrains when choosing a route. 
Having said that, even though ACE enables the rover to drive on significantly more difficult terrains than GESTALT, this conservatism is one of the remaining limitations.
Mitigating the conservatism of ACE on a highly undulating terrain is our future work.

\subsubsection{Simulation with multiple ACE runs}

We then drive a rover with a pre-specified path on various terrains in simulation while calling ACE multiple times at a fixed interval to check collisions.

\paragraph{Flat Terrain with a Bump}

\begin{figure}
    \centering
    \begin{subfigure}{0.32\textwidth}
        \centering
        \includegraphics[width=\textwidth]{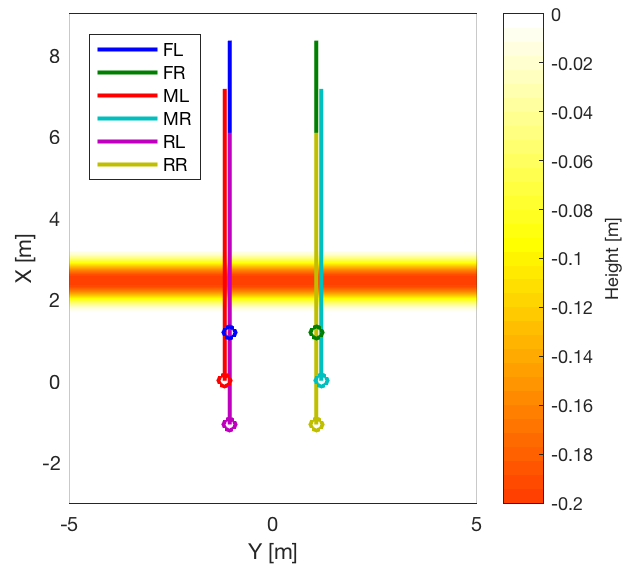}
        \caption{}
    \end{subfigure}
    \begin{subfigure}{0.32\textwidth}
        \centering
        \includegraphics[width=\textwidth]{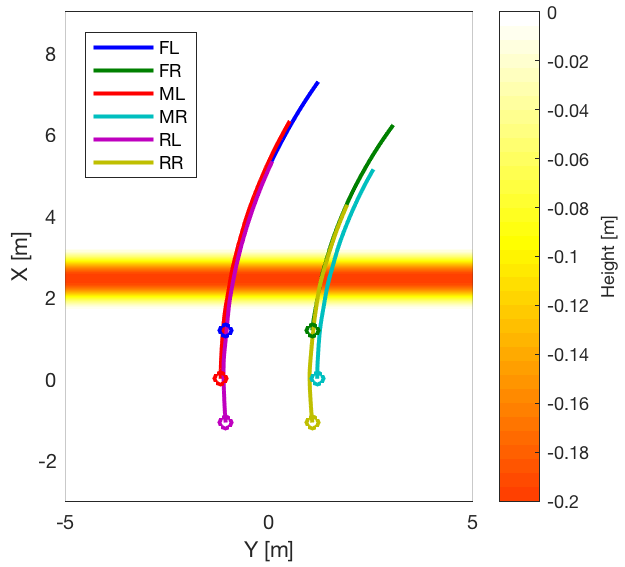}
        \caption{}
    \end{subfigure}
    \begin{subfigure}{0.32\textwidth}
        \centering
        \includegraphics[width=\textwidth]{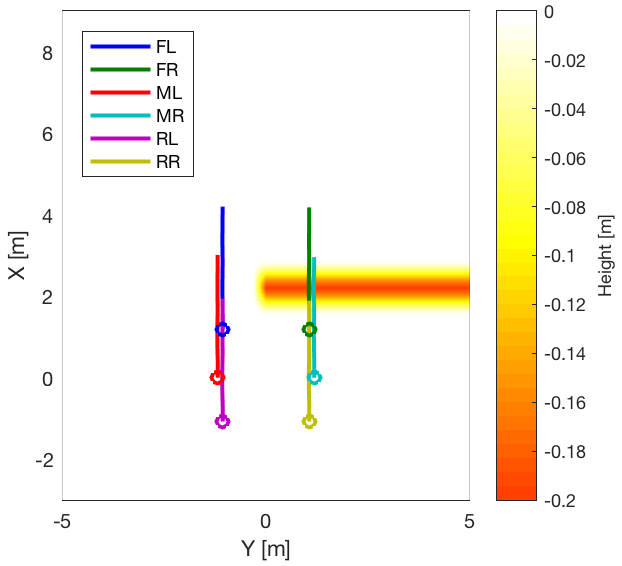}
        \caption{}
    \end{subfigure}
    \caption{Flat terrain with a smooth bump. Six wheel trajectories are shows in different color. The initial positions of the wheels are marked with circles. (a) Linear approach to the bump. (b) Curved approach to the bump. (c) Climb over the bump only with the right wheels.}
    \label{fig:wheel_tracks}
\vspace{14px}
    \centering
    \begin{subfigure}{\textwidth}
        \centering
        \includegraphics[width=\textwidth]{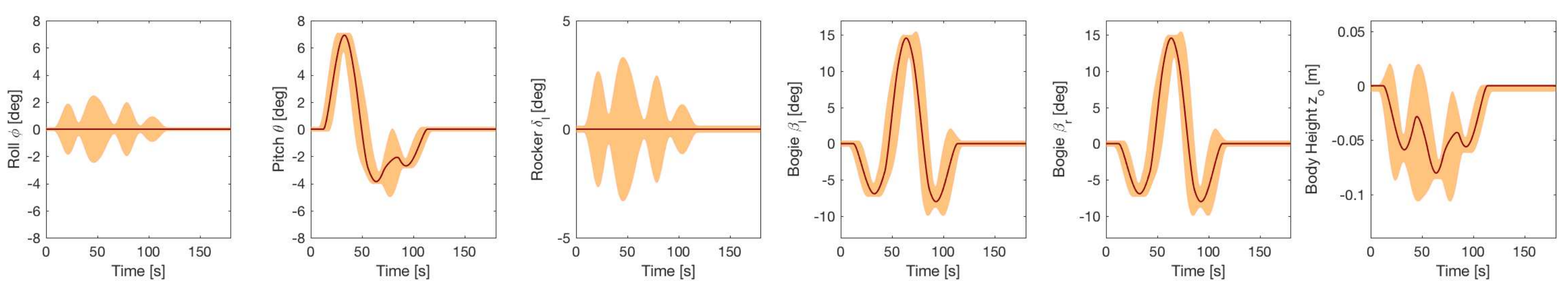}
        \caption{Linear motion}
    \end{subfigure}
    \begin{subfigure}{\textwidth}
        \centering
        \includegraphics[width=\textwidth]{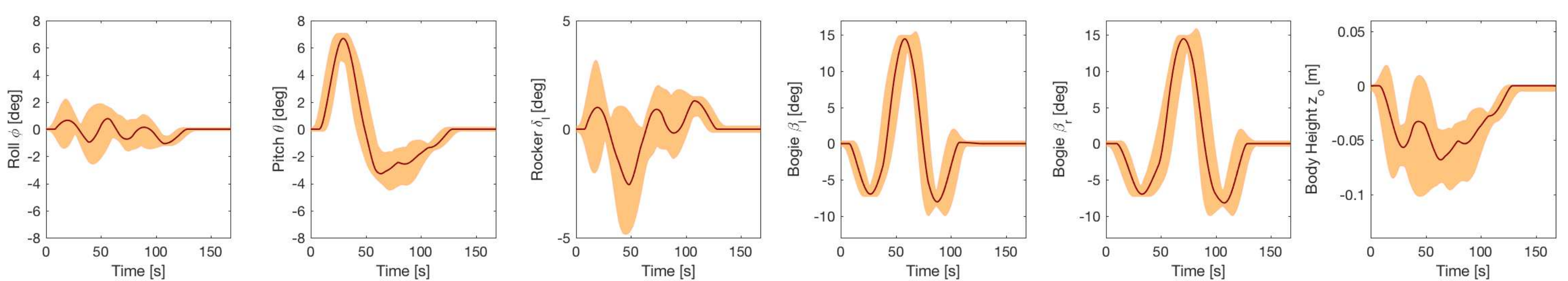}
        \caption{Curved motion}
    \end{subfigure}
    \begin{subfigure}{\textwidth}
        \centering
        \includegraphics[width=\textwidth]{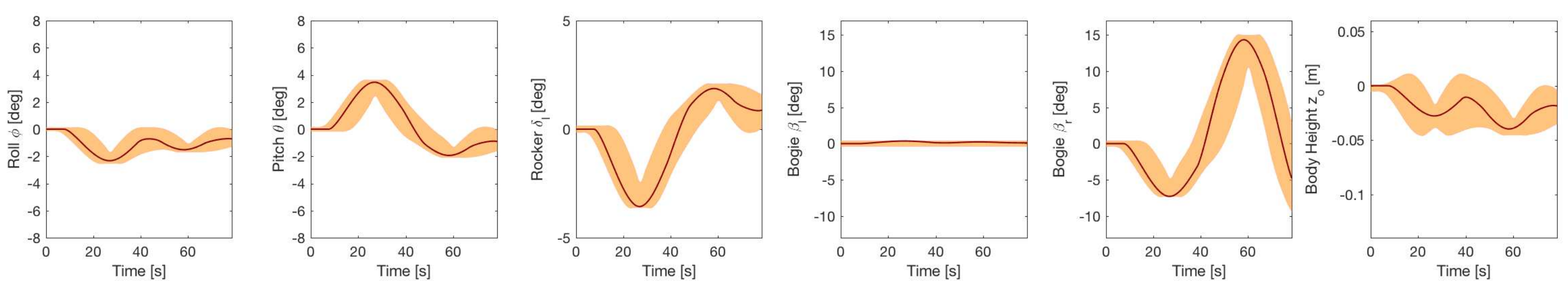}
        \caption{Linear motion with the right wheels on the bump}
    \end{subfigure}
    \caption{ACE estimation results for body roll $ \phi $, body pitch $ \theta $, left rocker angle $ \delta_l $, left and right bogie angles $ \beta_l $ and $ \beta_r $, and body height $ z_o $. The solid lines represent the ground-truth state computed by a numeric method. The shaded regions represent the ranges between the \ac{ACE} upper/lower bounds. }
    \label{fig:bump_result}
\end{figure}

The test environment is a simple flat terrain with a 0.2\,[m] height bump. A  Curiosity-sized rover is driven over the bump with three different trajectories shown in \autoref{fig:wheel_tracks}. The rover drives at the nominal speed of Curiosity on Mars ($\sim$0.04\,[m/s]). We collected data in 8\,[Hz], including ground-truth pose from a numeric method. The ranges of six wheel heights are extracted directly from the base map using the ground-truth pose reported by the simulator. 

\autoref{fig:bump_result} shows the time-series profiles of suspension and body states for three trajectories. The solid lines denote the ground-truth states computed by the numeric method, and the shaded regions represent the state bounds computed by \ac{ACE}. 
All the ground-truth states are always within the bounds, meaning ACE bounds are conservative as expected.  
It is interesting to observe how the algorithm evaluates rover states for its worst-case configurations. With trajectory (a), the rover approaches perpendicularly to the bump. The ground-truth roll angle stays zero for the entire trajectory since the left and right wheels interact with the ground exactly at the same time in this noise-free simulation. However, this is unlikely in the real-world settings where small difference in contact time, or difference in wheel frictions, can disturb the symmetry and cause rolling motion. \ac{ACE} computes the state bounds based on the worst-case configurations. Therefore, the algorithm captures such potential perturbation and conservatively estimate the state bounds, as presented in the top left figure of \autoref{fig:bump_result}.

\paragraph{Martian Terrain Simulation}

\begin{figure}
    \centering
    \begin{subfigure}[b]{0.4\textwidth}
        \centering
        \includegraphics[width=0.8\textwidth]{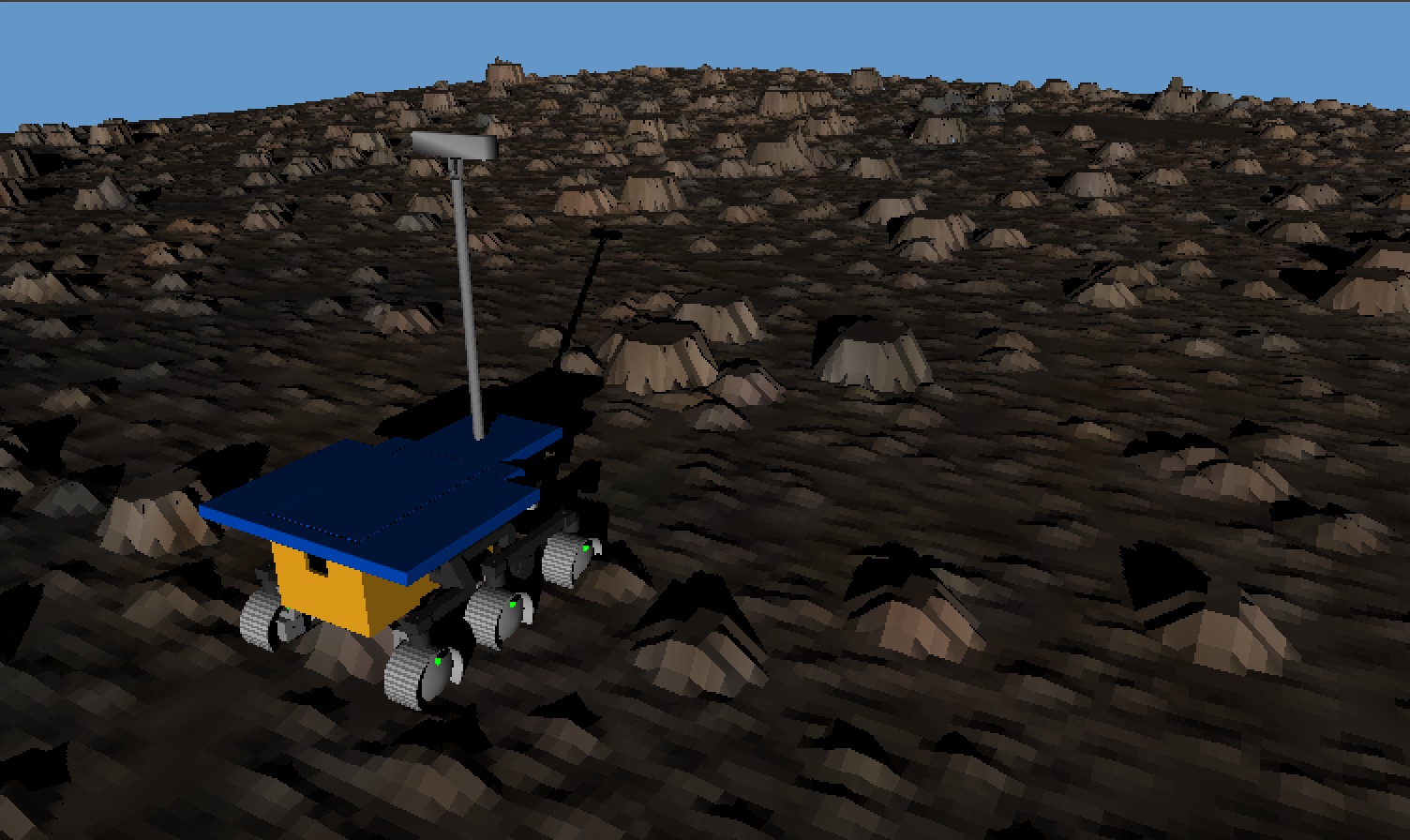}
        \caption{Simulator view}
        \includegraphics[width=0.9\textwidth]{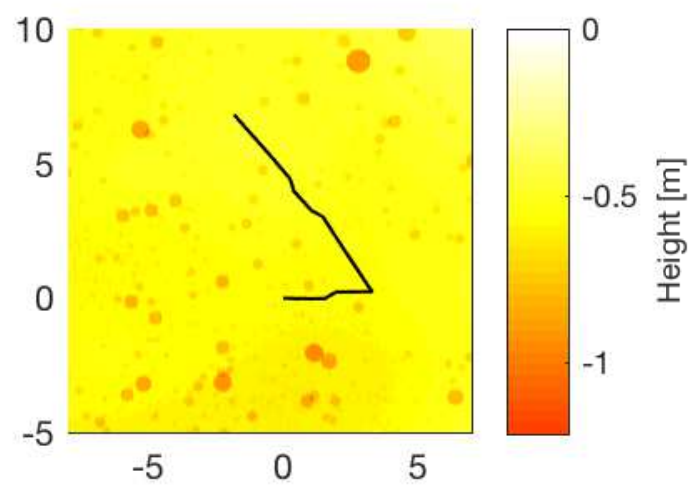}
        \caption{Path on \ac{DEM}}
    \end{subfigure}
    \begin{subfigure}[b]{0.58\textwidth}
        \centering
        \includegraphics[width=\textwidth]{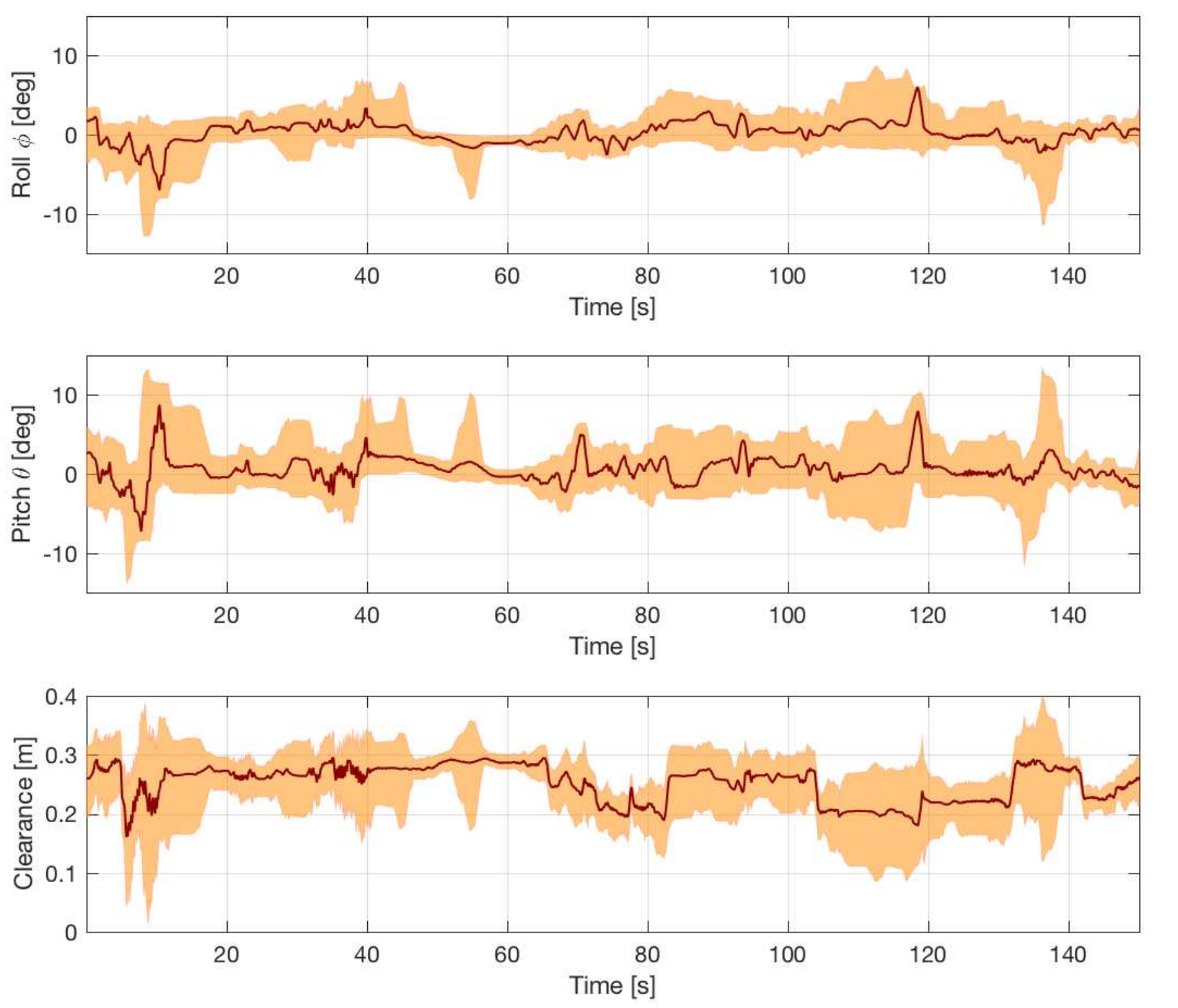}
        \caption{State estimation result}
    \end{subfigure}
    \caption{ACE estimation result in a Mars-like environment. The Rocky 8 rover was deployed on a synthetic Martian terrain. The base terrain is an orbit-based \ac{DEM} for the Jezero crater. The rocks are randomly populated according to the Martian size-frequency distribution model.}
    \label{fig:roams_result}
\end{figure}

Next, we tested the \ac{ACE} algorithm in a Mars-like environment. We imported a \ac{DEM} of Jezero crater into the \ac{ROAMS} simulator \cite{Jain2003}. Since no spacecrafts have landed on Jezero, we only have a limited resolution of terrain model from satellite measurements. To create an environment closer to the actual, we populate rocks based on the empirically created Martian rock size-distribution model \cite{Golombek2008_CFA}. We populate rocks assuming 10\% \ac{CFA}. For the hardware platform, we used the Rocky 8 rover which is a mid-sized rover similar to \acp{MER}. We drove the rover at a speed of 0.15\,[m/s] with an autonomous hazard avoidance mode. The data are taken at 10\,[Hz] including ground-truth pose reported by the simulator.

The state estimation result is shown in \autoref{fig:roams_result}. The figure only reports the body states including roll, pitch, and minimum clearance, but similar results are obtained for the other suspension states. Again, all the ground-truth states are always within the \ac{ACE}  bounds, successfully confirming the algorithmic conservatism.
The level of conservatism varies from time to time. 
For most of the time, the span between upper and lower bounds were within a few degrees. 
However, at 55\,[s] in \autoref{fig:roams_result}, for example, the upper bound on the pitch angle was about 10\,[deg] while the actual angle is around 1\,[deg]. 
Such false alarms typically occur when a large rock is in one of the wheel boxes but the rover did not actually step on it.
This behavior is actually beneficial because it helps the path planner to choose less risky paths if the planner uses the bounds as a part of its cost function. 
Of course, such conservatism may result in a failure of finding a feasible path. 
However, we reiterate that conservatism is an objective of ACE because safety is of supreme importance for Mars rovers. 
Furthermore, as we will demonstrate in \autoref{sec:result_conservatism}, ACE significantly reduces the conservatism compared to the state-of-the-art. 
An additional idea that can further mitigate the conservatism is to introduce a probabilistic assessment, as proposed by \cite{Ghosh2018}.

\subsection{Hardware Experiments}
\label{sec:experiments}
We deployed ACE on actual hardware systems and validated through the field test campaign in the JPL Mars Yard. ACE is deployed on two rover testbeds: the Athena rover whose size is compatible to the MER rovers, and the Scarecrow rover, a mobility testbed for MSL. In both systems, terrain height measurements are obtained by the stereo camera attached to the mast.
Therefore, the heightmap that ACE receives involves noise from the camera and stereo processing. As we will report shortly, the stereo noise results in occasional bound violations. 
Adding an adequate margin to account for the noise restores the conservatism of ACE. 

\subsubsection{Athena Rover}

\begin{figure}
    \centering
    \includegraphics[width=0.8\textwidth]{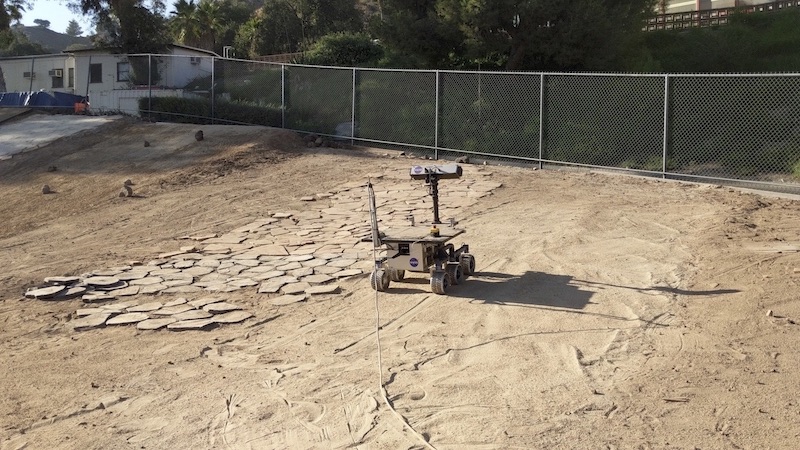}
    \caption{Athena rover driving on a slope of JPL Mars Yard.}
    \label{fig:athena}
\end{figure}

\begin{figure}
    \centering
    \begin{subfigure}[b]{0.48\textwidth}
        \centering
        \includegraphics[width=\textwidth]{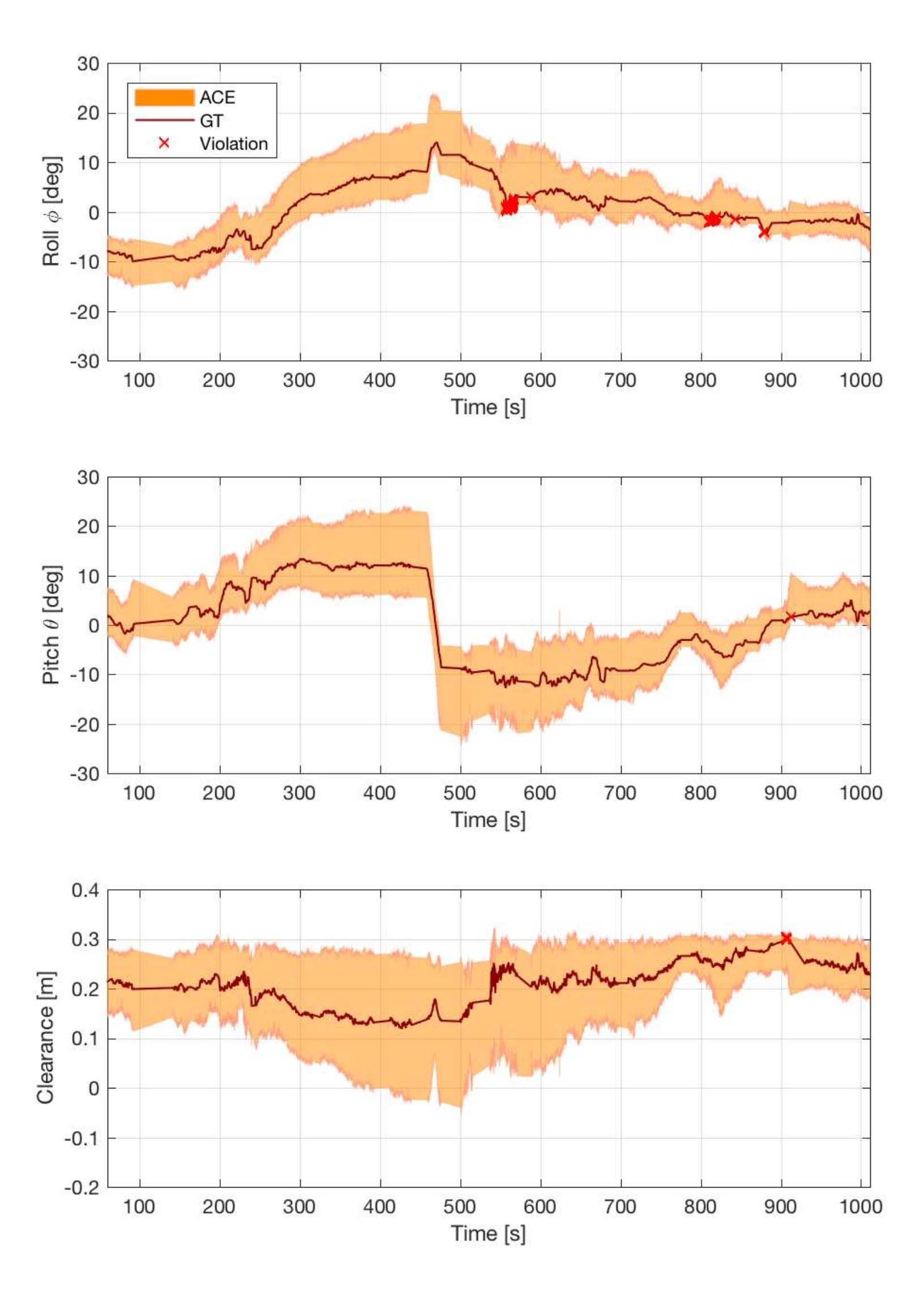}
        \caption{}
    \end{subfigure} 
    \begin{subfigure}[b]{0.48\textwidth}
        \centering
        \includegraphics[width=\textwidth]{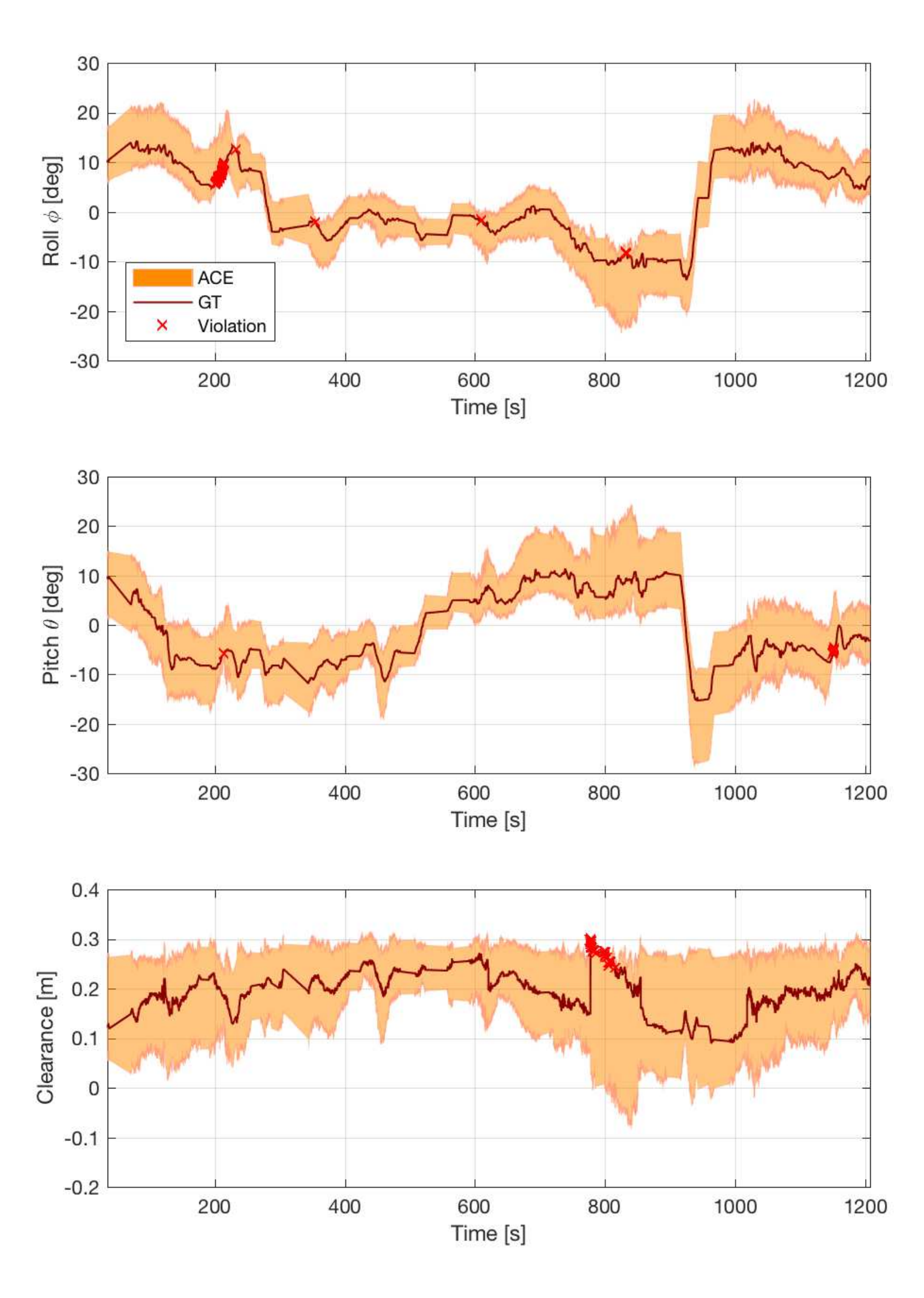}
        \caption{}
    \end{subfigure} 
    \caption{ACE estimation result for Athena's two different drives. The ground-truth (GT) state is within the ACE bounds except occasional violations.}
    \label{fig:athena_result}
\end{figure}

The first experiment is performed with the Athena rover developed at \ac{JPL} (see \autoref{fig:athena}). The platform is designed for testing Mars rover capabilities on Earth and is comparable in size to \ac{MER}. The navigation system is primarily vision-based using a stereo-camera pair consisting of two PointGrey Flea2 cameras rigidly mounted on a movable mast. The mast is at a height of 1.4\,[m], and the baseline of the stereo-camera pair is 0.30\,[m]. The images are captured at a resolution of $640 \times 480$ from wide field-of-view lenses. The ground-truth pose is obtained from OxTS xNAV system, which reports 6DoF pose from integrated GPS and inertial measurements. The suspension angles are not measured on this platform.

We manually drove Athena on a slope of 6 to 12\,[deg] in the Mars Yard. The slope consists of multiple terrain materials including cohesive/cohesion-less sands and bedrocks. We evaluated \ac{ACE} by comparing the estimated state bounds from the algorithm and the ground-truth state. \ac{ACE} was applied to a past few image sets prior to the driving time. Unlike the rover autonomous drive software that prevents the placement of wheels in unknown terrain, our dataset collected by manual commanding contains samples in which the point clouds do not cover the terrain under all six wheels. We do not report state estimation results for such incomplete data.

\autoref{fig:athena_result} shows the ground truth, as well as the upper and lower bounds from ACE, of the roll and pitch angles and the ground clearance for two drives. 
Each run consists of about 40\,[m] traverse including level, uphill, downhill, and cross-slope drives. 
As expected, the ground truth is within the bounds for most of the time. 
However, unlike the simulation results reported in the previous subsection, occasional bound violations were observed, as shown by the red crosses on the plots. 
This was due to perception errors, such as stereo matching error and calibration error. The positional error in point clouds from the stereo camera is propagated to the rover states through the kinematic equations, causing the error in state bounds. 
A practical approach to restore the conservatism is to add a small margin $\epsilon$ to the perceived height of the ground. 
More specifically, the maximum and minimum height of each wheel box, $z^+_w$ and $z^-_w$ ($w \in \{ f, m, r\}$) in (\ref{eq:21})-(\ref{eq:32}), are replaced with $z^+_w + \epsilon$ and $z^-_w - \epsilon$, where $\epsilon$ is the estimated upper bound of the height error.
A downside of this approach is an increased conservatism.

\autoref{tab:athenta_stats} shows a statistical result from cumulative 130\,[m] drive by Athena. The total success rate is computed by counting samples that all state variables are within the \ac{ACE} bounds. The success rate was 98.74\% without the perception error margin, with $\sim 3$\,[deg] maximum attitude error and 0.012\,[m] clearance error. Although it is rare that these small estimation error contributes to the hazard detection miss which is critical to the mission, extra conservatism is preferred for planetary applications. The conservatism is fully restored (i.e., 100\% success rate) with $\epsilon=15$\,[mm], which roughly corresponds with the worst-case height perception error.

\begin{table}[t]
    \centering
    \caption{Error statistics from cumulative 130\,[m] drive by Athena.}
    \label{tab:athenta_stats}
    \begin{tabular}{lrrrr}
        \toprule
         & & \multicolumn{3}{c}{Max State Violation} \\
        \cmidrule(lr){3-5}
        Method & Success Rate [\%] & Roll [deg] & Pitch [deg] & Clearance [m] \\
        \midrule
        ACE & 98.74 & 2.6 & 3.2 & 0.012 \\
        ACE ($\epsilon$=5\,[mm]) & 99.70 & 1.7 & 2.0 & 0 \\
        ACE ($\epsilon$=10\,[mm]) & 99.95 & 0.7 & 0.9 & 0 \\
        ACE ($\epsilon$=15\,[mm]) & 100 & 0 & 0 & 0 \\
        \bottomrule
    \end{tabular}
\end{table}

\subsubsection{Scarecrow Rover}

\begin{figure}
    \centering
    \includegraphics[width=0.7\textwidth]{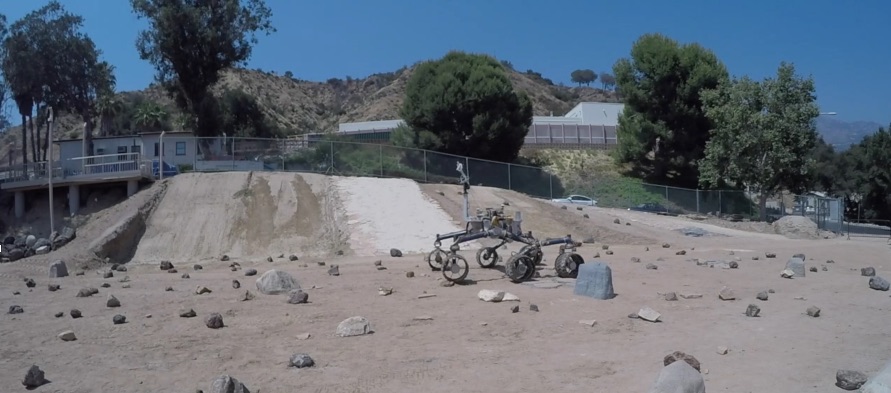}
    \caption{Scarecrow test in JPL's Mars Yard on July 17, 2018, showing a typical set-up of the Yard for the experiments.}
    \label{fig:marsyard_for_scarecrow}
\end{figure}

We deployed ACE on JPL's mobility testbed called Scarecrow and performed a series of experiments in JPL's Mars Yard. The purpose of the experiments is to test ACE with hardware and software that is close to the Mars 2020 rover. The mobility hardware of Scarecrow, including the rocker-bogie suspension system and wheels, are designed to be nearly identical to that of Curiosity and Mars 2020 rovers. The vehicle's mass is about one third of Curiosity and Mars 2020 rovers, simulating their weight under the Martian gravity. In terms of software, ACE is re-implemented in C and integrated with the Mars 2020 flight software. Since Scarecrow was originally designed for mobility experiments, it does not have the identical processor as the real Mars rovers. Instead, we compiled the software for Linux and run on a laptop computer placed on the top deck of the vehicle. Therefore, this experiment does not replicate the run time of the software. We evaluated the run time of ACE in a hardware-in-the-loop simulation using RAD750, as described in Section \ref{sec:runtime}. The original design of Scarecrow also lacks cameras. Therefore, we retrofitted a pair of Baumer cameras, from which height map is created on-the-fly via on-board stereo processing. The Mars Yard is configured in a way to represent some of the most difficult conditions in the Mars 2020 landing site, including 30 degree of slope and 15\% \ac{CFA} \cite{Golombek2008_CFA} of rock density. \autoref{fig:marsyard_for_scarecrow} shows a typical set up of the Mars Yard.

Our extensive test campaign consisted of 42 days of experiments in the Mars Yard using Scarecrow. The analysis of the test results were largely qualitative rather than quantitative or statistical for a few reasons. First, we are unable to keep the exactly same set up of the Mars Yard as it is shared by many teams. It is also slightly altered by precipitation and wind. Second, the driving speed of Scarecrow is only 0.04\,[m/s] (same as Curiosity and Mars 2020 rover), therefore it typically takes 20 to 30 minutes to complete a single Mars Yard run. Repeating a statistically significant number of runs with the same set up is difficult. Third, the software implementation was continuously improved throughout the test campaign. Fourth, the ground truth of belly pan clearance is difficult to measure directly. Fifth and finally, the tests were performed as a part of the software development for the Mars 2020 rover mission, where the main purpose of the tests were the verification and validation of the integrated software capabilities rather than the quantitative assessment of the performance of ACE alone.  

\begin{figure}
\centering
    \includegraphics[width=0.7\textwidth]{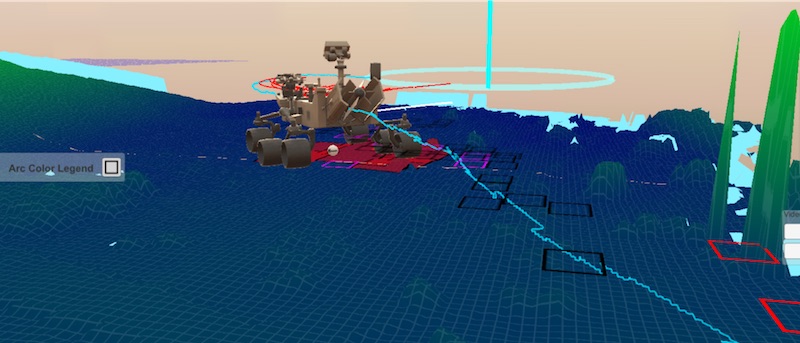}
\caption{Visualization of a path planned in JPL's Mars Yard on July 23, 2018 with \textit{Caspian} visualizer.}
    \label{fig:caspian}
\end{figure}

Qualitatively, through the test campaign, the algorithm and implementation were matured to the point where the vehicle can drive confidently over $ \sim40 $\,[m] through a high rock density (15\% CFA) terrain. Since the path planner solely rely on ACE for collision check, the fact that the rover reliably avoids obstacles without hitting the belly pan is an indirect and qualitative evidence that ACE is working properly. For example, \autoref{fig:caspian} shows the 3D reconstruction of the terrain and the vehicle configuration from the Scarecrow test data. 

\begin{figure}
    \centering
    \begin{subfigure}{\textwidth}
        \centering
        \includegraphics[width=\textwidth]{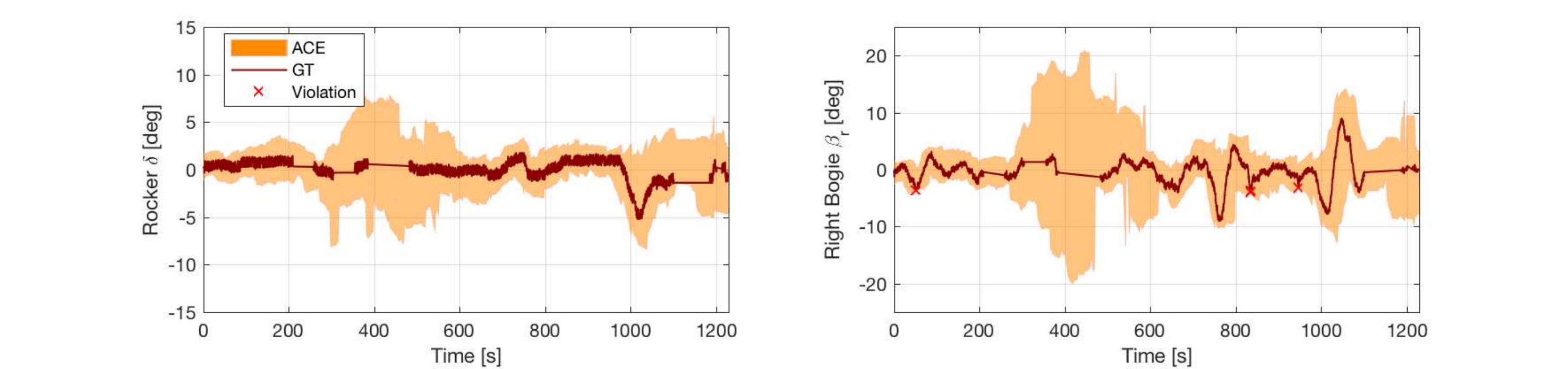}
        \caption{}
    \end{subfigure}
    \begin{subfigure}{\textwidth}
        \centering
        \includegraphics[width=\textwidth]{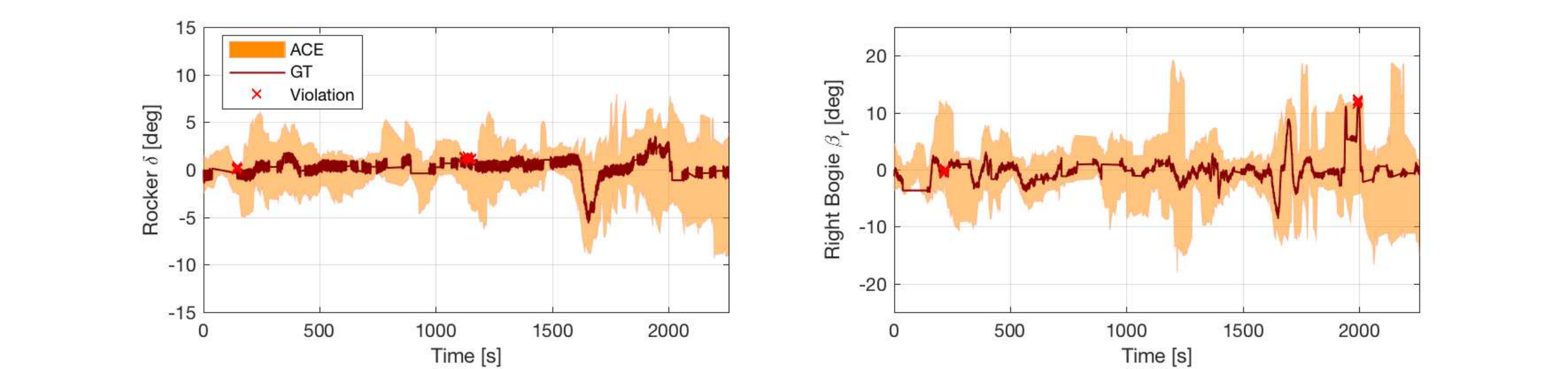}
        \caption{}
    \end{subfigure}
    \begin{subfigure}{\textwidth}
        \centering
        \includegraphics[width=\textwidth]{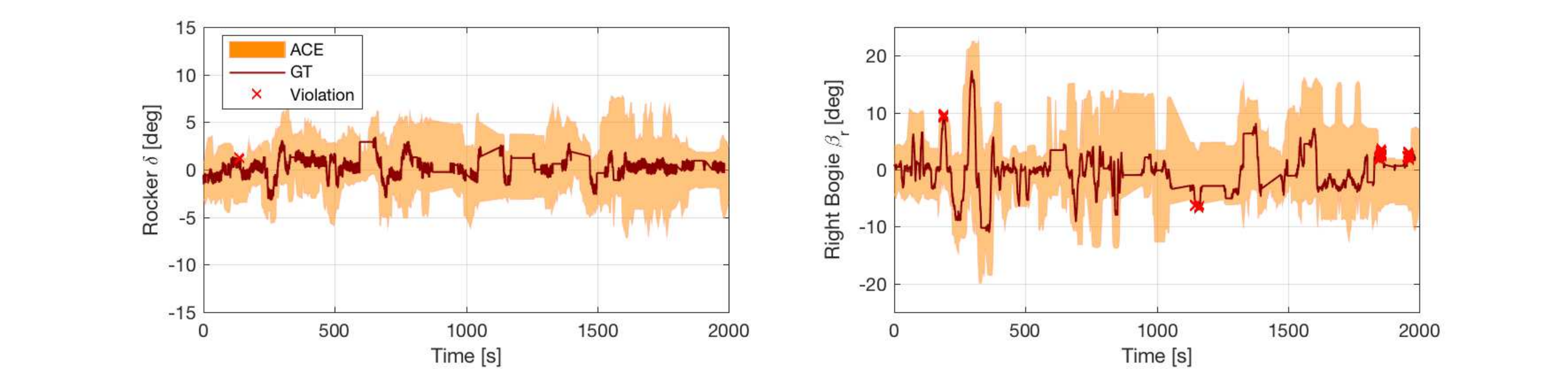}
        \caption{}
    \end{subfigure}
\caption{Recorded ground-truth (GT) rocker and right bogies angles as well as the predicted bounds by ACE on three Scarecrow tests performed on Sept 12, 2018, in JPL's Mars Yard.}
    \label{fig:scarecrow_results}
\end{figure}

A limited quantitative assessment is possible because a few intermediate and derivative variables in ACE were directly measured and recorded. These variables include rocker angle, left and right bogie angles, and the vehicle's tilt angle. Figure \ref{fig:scarecrow_results} shows the ground-truth measurement of rocker and right bogie angles as well as the bounds computed by ACE on three long Scarecrow drives in the Mars Yard. There are a few observations from the results. Firstly, the bounds successfully captured the ground-truth trends. For example, the negative spike in the rocker angle at $\sim 1100$\,[s] in Figure \ref{fig:scarecrow_results}(a) is correctly predicted by ACE, indicated by the reduced lower bound around that time. Secondly, the bounds were almost always respected. Thirdly, however, we observed occasional violations of the bounds as shown in red crosses on the plots. Our investigation concluded that the main causes of bound violations are the error in encoders and the error in perceived height map. The height map error is a result of two factors: 1) error in stereo processing (i.e., feature extraction and matching, error in camera model, noise in images, etc) and 2) ``smoothing effect" due to re-sampling (3D point cloud from stereo processing is binned and averaged over a 2D grid). This conclusion was derived by using simulations in the following steps: 1) assured that ACE bounds are always respected when running ACE on a ground-truth height map, 2) reproduced the stereo error by using simulated camera images, and 3) ACE bound violations occur with comparable frequency and magnitude with the simulated stereo error. 
As in the Athena rover experiment, adding an adequate margin $\epsilon$ on the perceived height can restore the conservatism.

\subsubsection{Run-time Performance}\label{sec:runtime}

The run-time performance is important for space applications where the computational resources are severely limited. \ac{ACE} has a significant advantage on this regard, compared to other alternatives that depend on iterative numeric methods. 
In the following analysis, we chose plane fitting as a point of comparison because it is a light-weight approximations for estimating rover state on rough terrain and used as the basis of GESTALT, the state-of-the-art autonomous navigation algorithm being used for the existing Mars rovers.

The computation of \ac{ACE} is very fast due to its closed-form formulation. On the NVIDIA Jetson TK1 board on the Athena rover, \ac{ACE} takes 11.2\,[$\mu$s] for a single pose evaluation while plane fitting takes 26.1\,[$\mu$s] over $\sim$100 points and 68.2\,[$\mu$s] over $\sim$200 points. \ac{ACE} runs faster than the naive plane-fit approach using least squares, as well as providing richer information about the vehicle state.
For reference, the average run-time of \ac{ACE} on a 2.8GHz Intel Core i7 machine is 2\,[$\mu$s], which enables a robot to evaluate 500k poses at a second, whereas plane-fit is 5 times slower with 200 points.
Next, perhaps more importantly for spacecraft applications, the computational time of \ac{ACE} is constant. Thanks to the analytic formulation of \ac{ACE}, the computational time is always the same regardless of terrain patterns. This is not the case for numeric methods, which require more iterations for complex terrain before converging.

We also evaluate the performance of ACE on the RAD750 CPU, which is used for the Curiosity and Mars 2020 rovers. While the precise timing is difficult due to the specialized configuration of the flight software, the typical run-time was 10-15 [ms] with a 10 [cm] resolution \ac{DEM}. This is sufficient run-time as a collision checker to support the ambitious traversal plans on the \ac{M2020} mission.

\subsection{Comparison with State-of-the-Art}
\label{sec:result_conservatism}

\begin{figure}
    \centering
    \begin{subfigure}{\textwidth}
        \centering
        \includegraphics[width=0.24\textwidth]{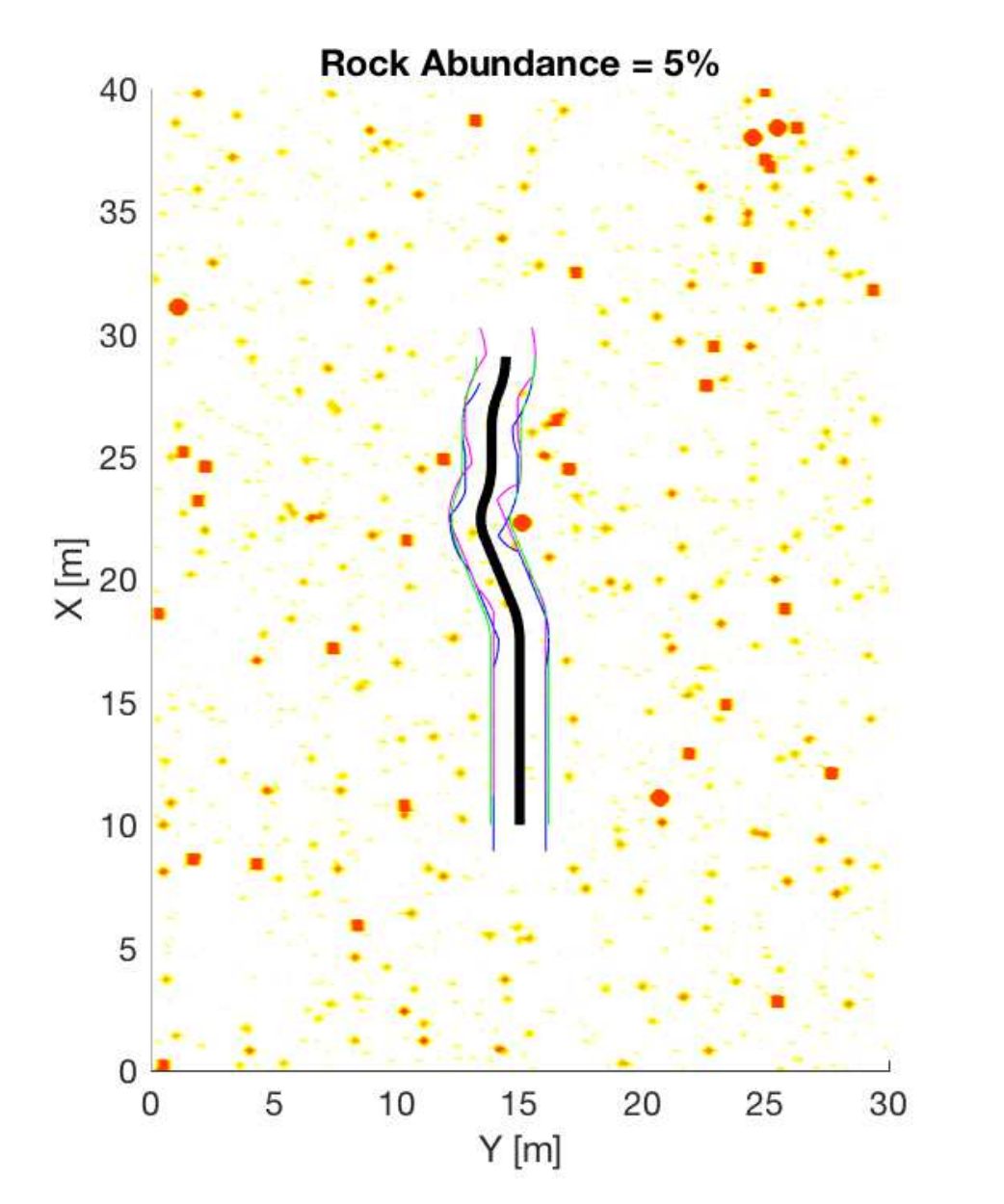}
        \includegraphics[width=0.24\textwidth]{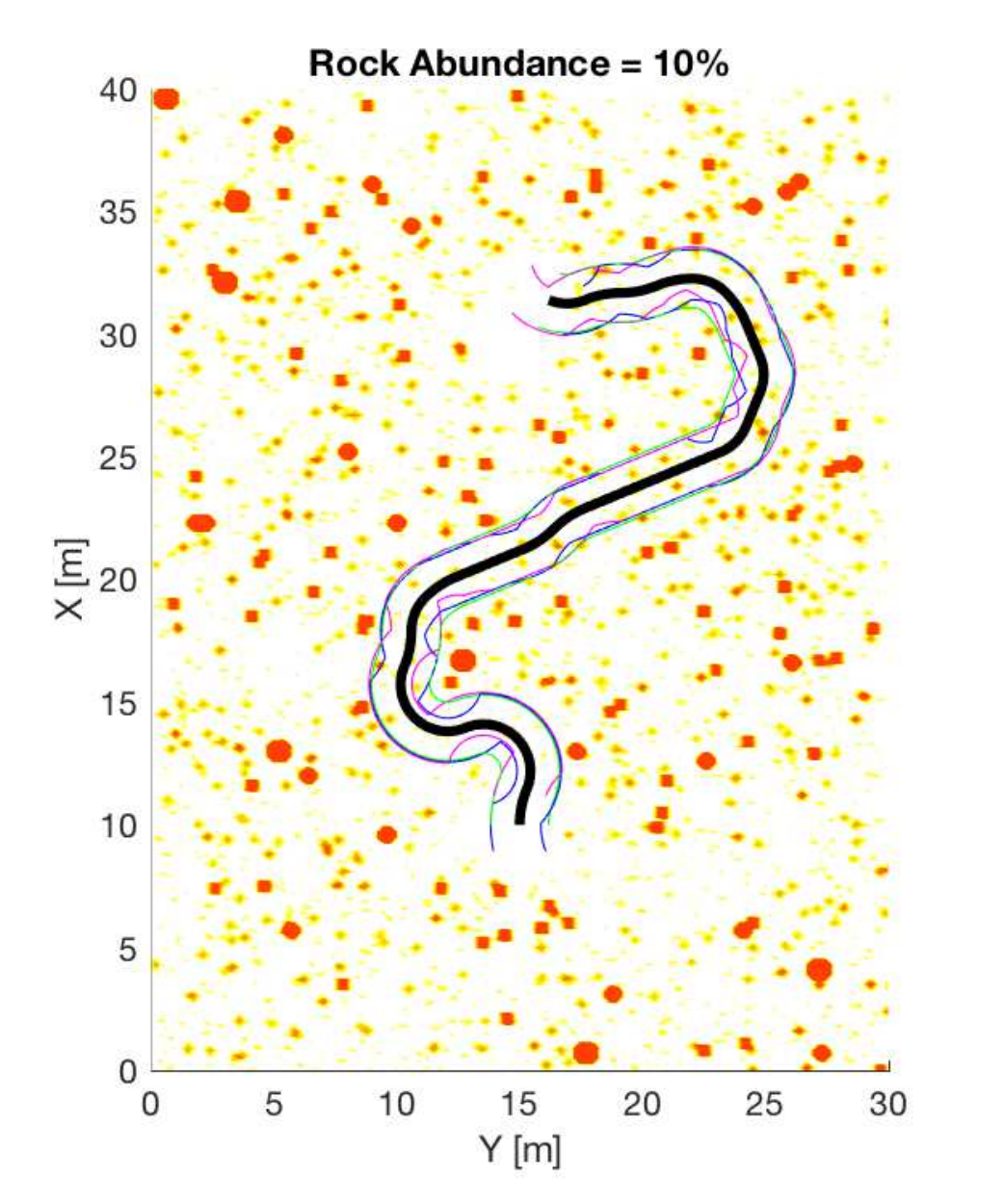}
        \includegraphics[width=0.24\textwidth]{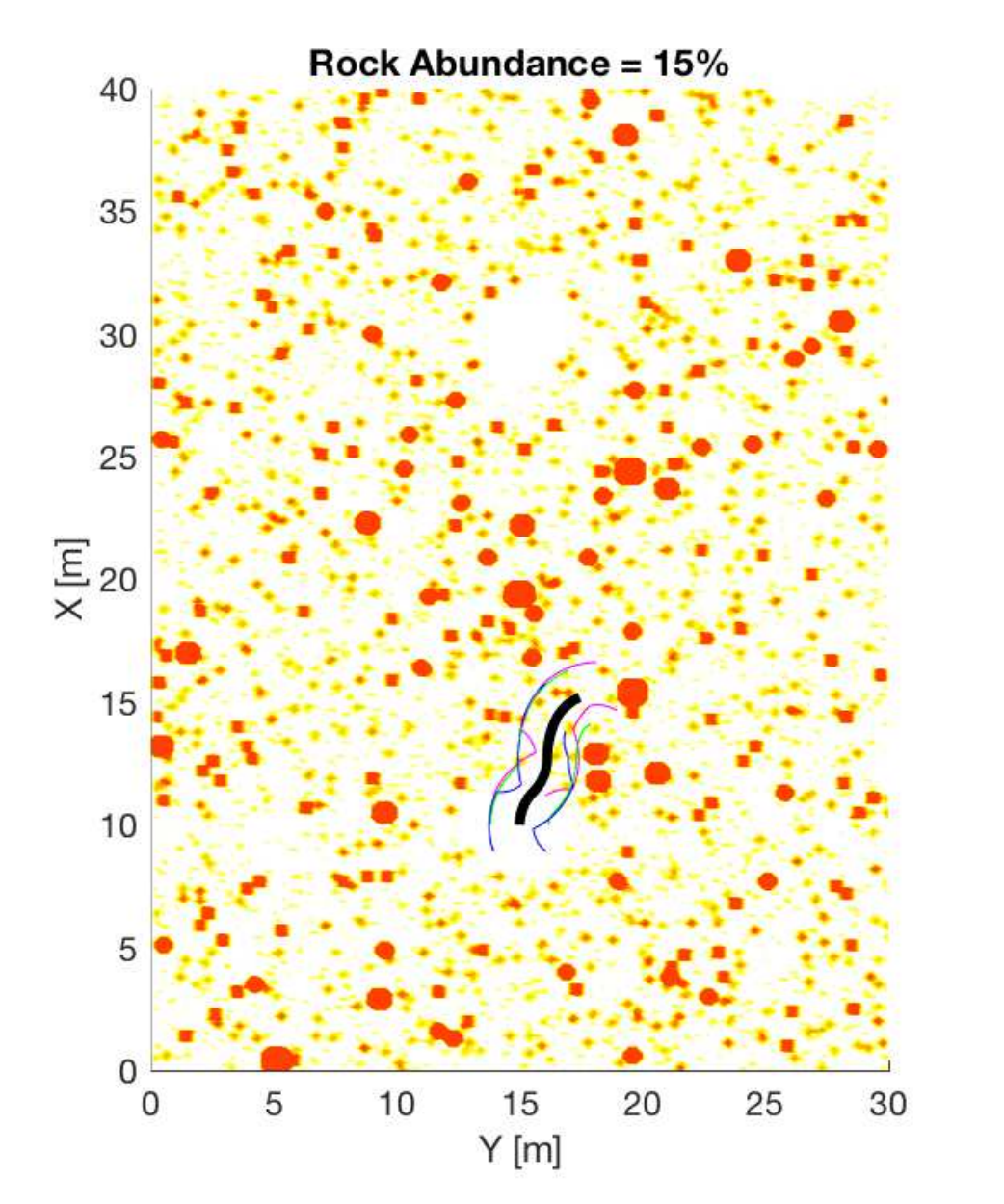}
        \includegraphics[width=0.24\textwidth]{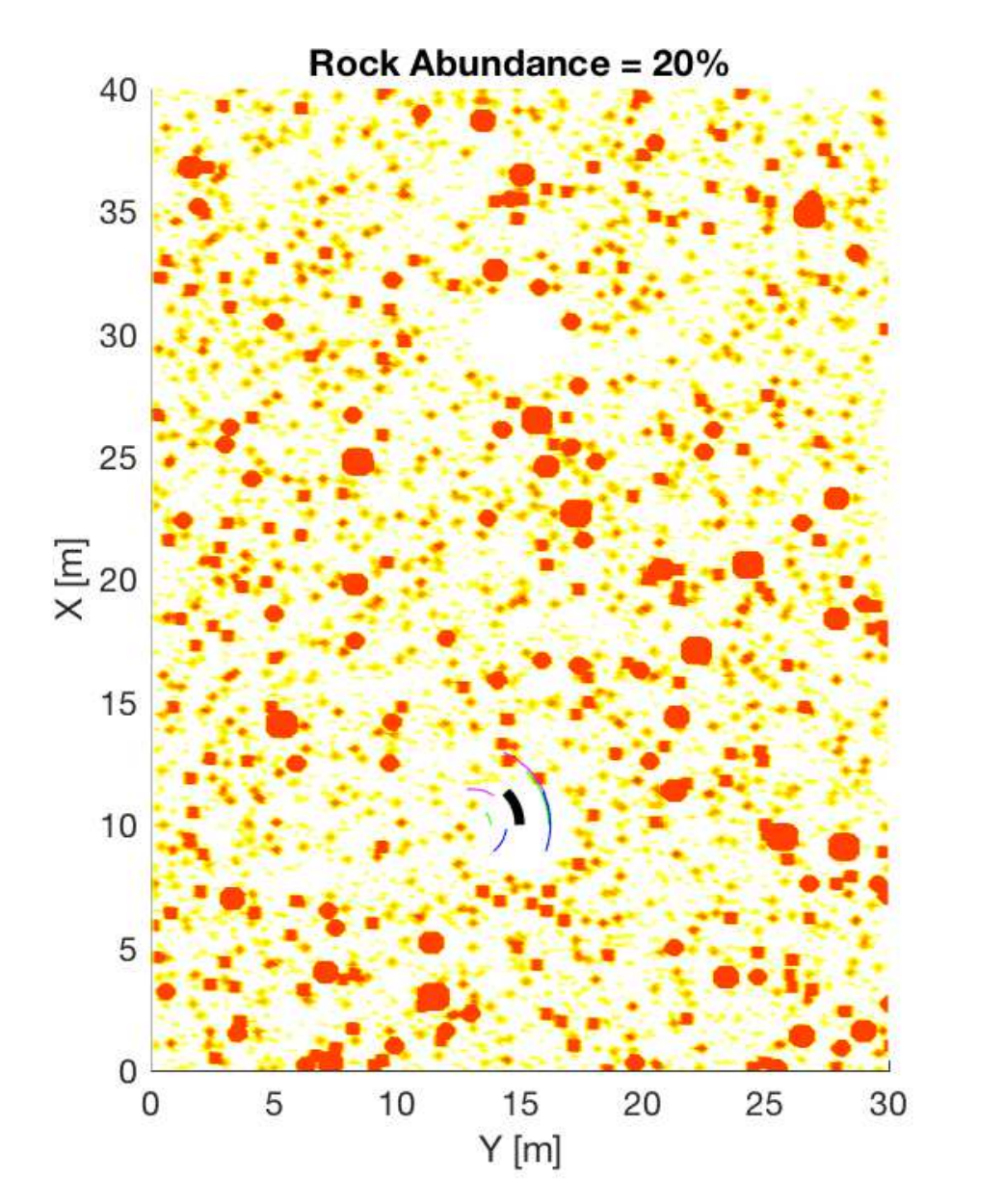}
        \caption{State-of-the-art path planner}
    \end{subfigure}
    \begin{subfigure}{\textwidth}
        \centering
        \includegraphics[width=0.24\textwidth]{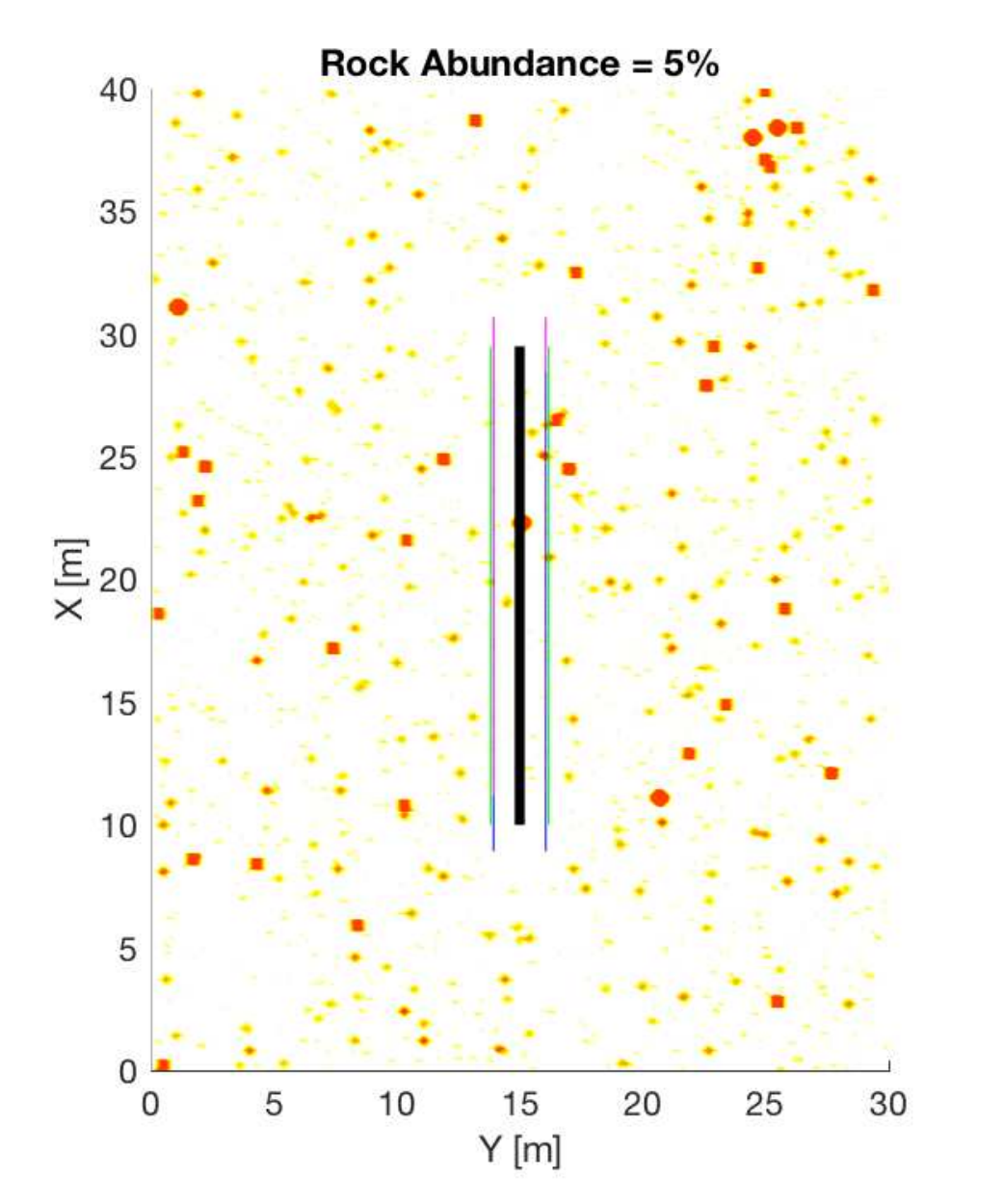}
        \includegraphics[width=0.24\textwidth]{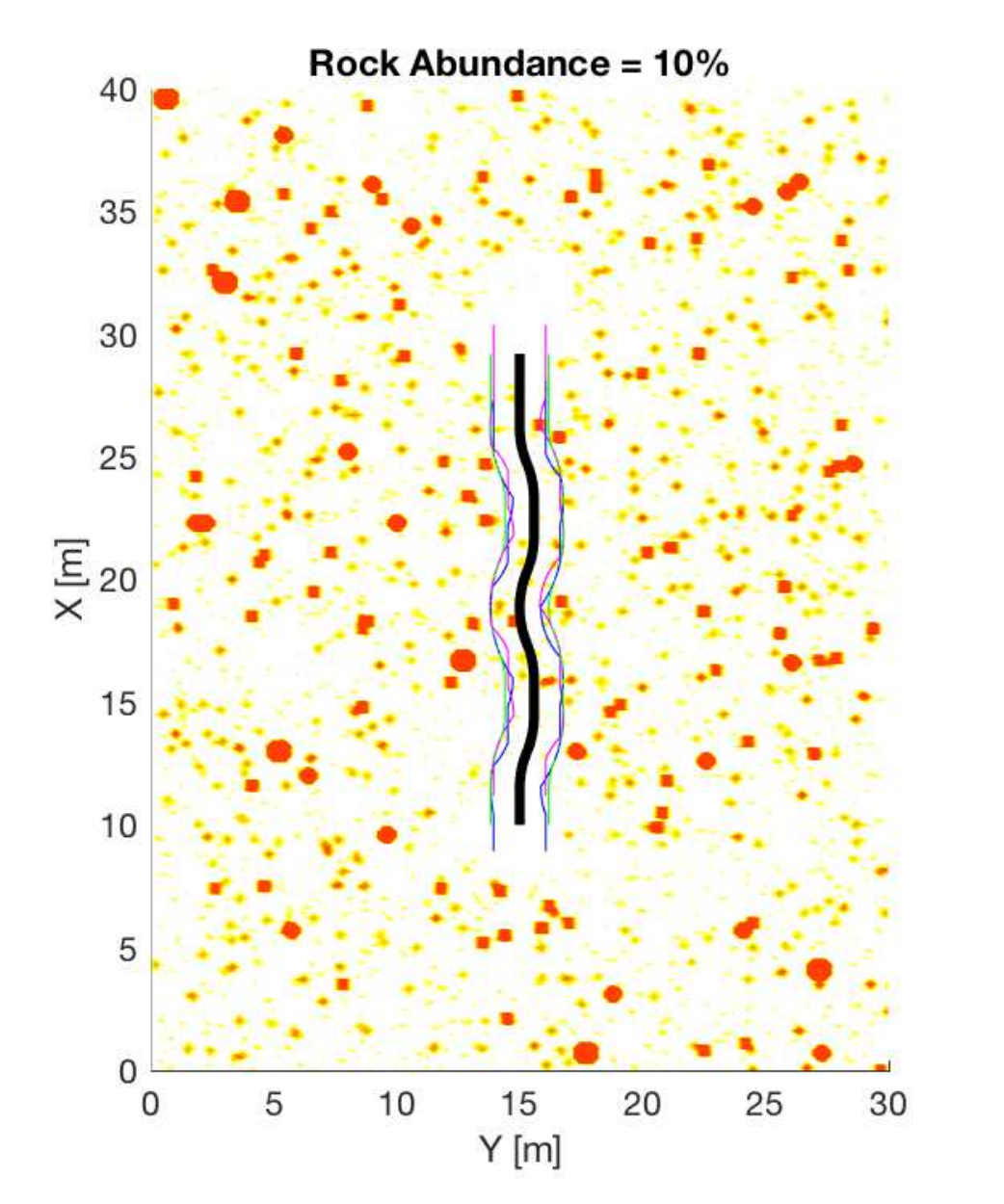}
        \includegraphics[width=0.24\textwidth]{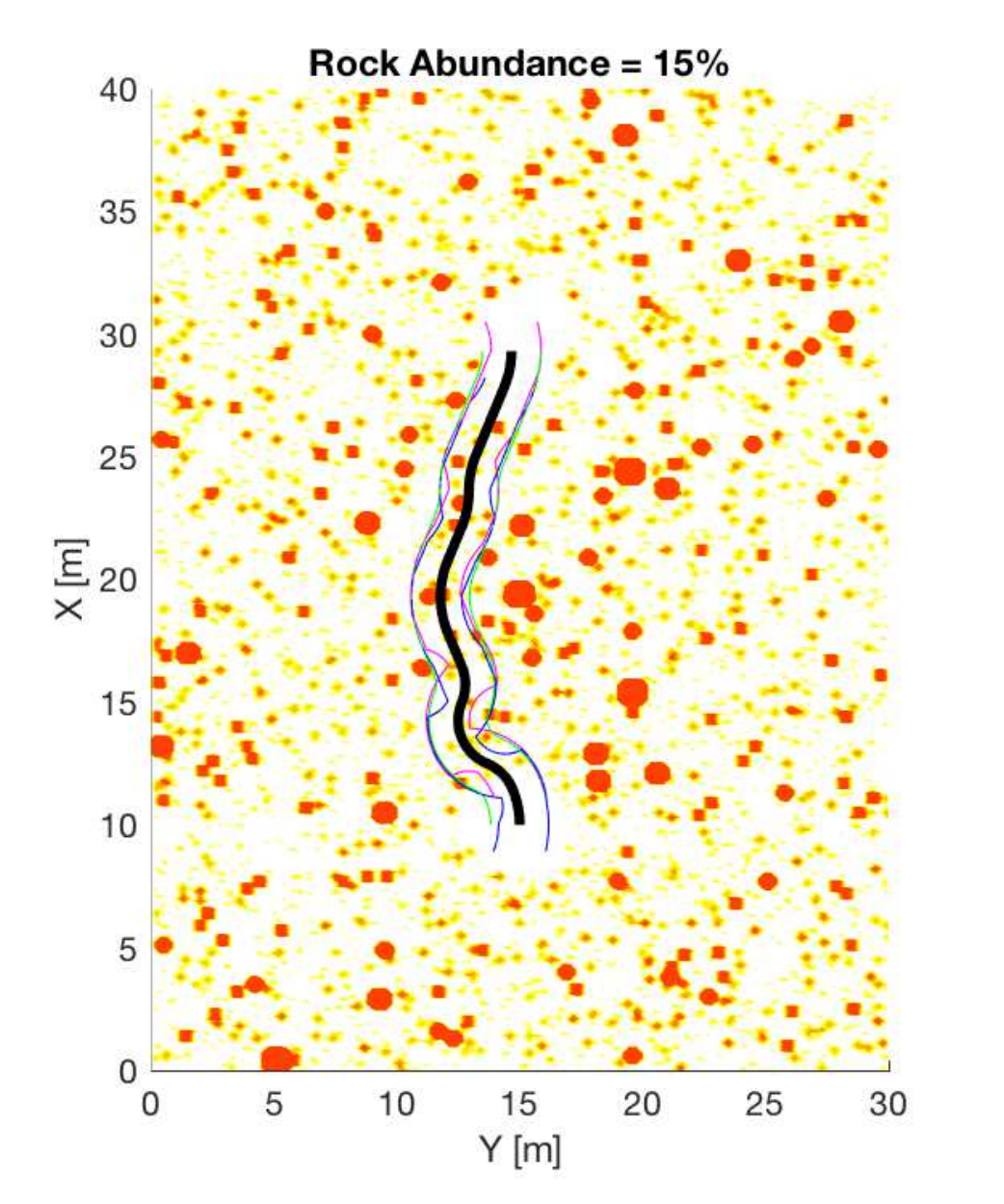}
        \includegraphics[width=0.24\textwidth]{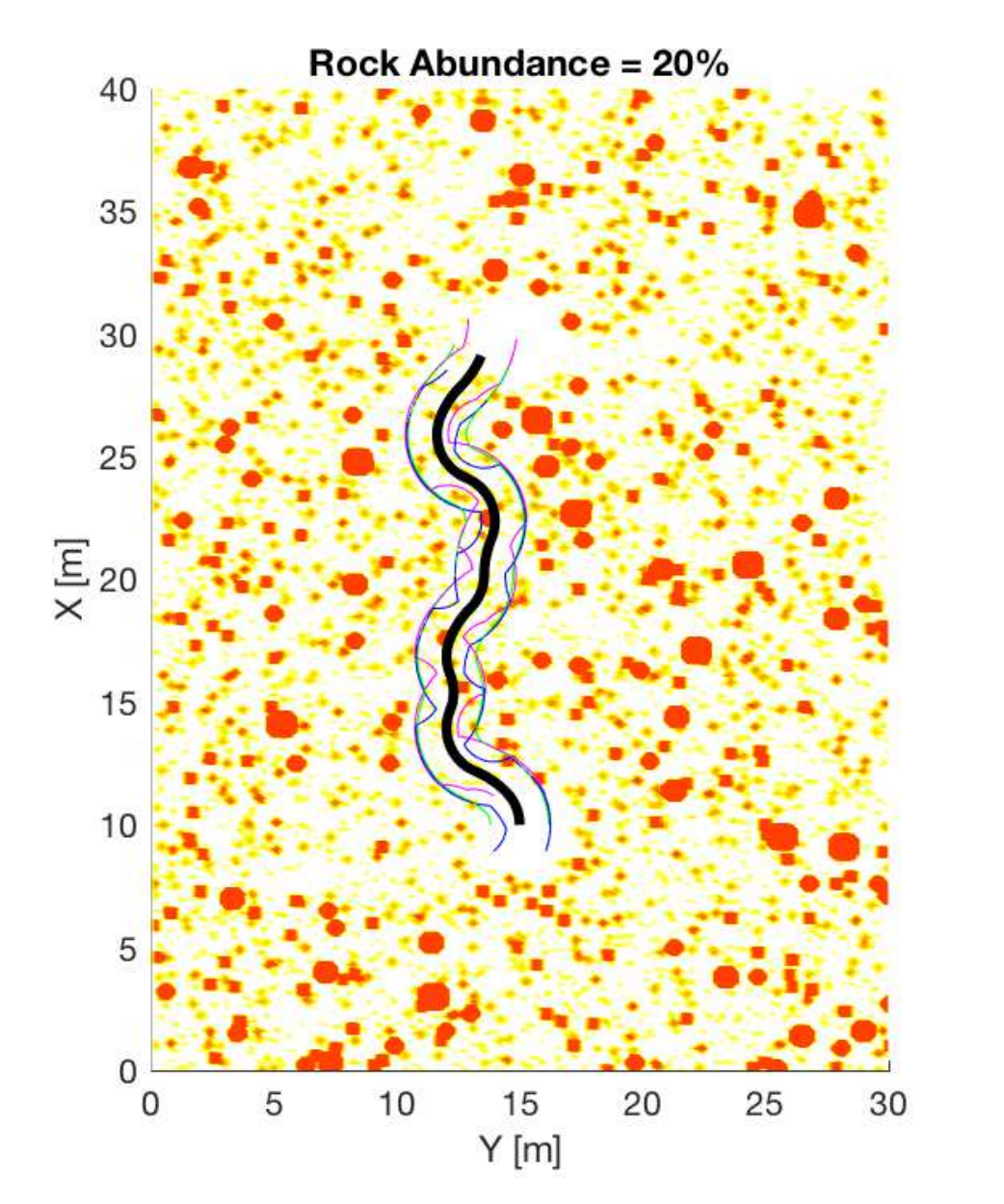}
        \caption{ACE-based path planner}
    \end{subfigure}
    \begin{subfigure}{\textwidth}
        \centering
        \includegraphics[width=0.24\textwidth]{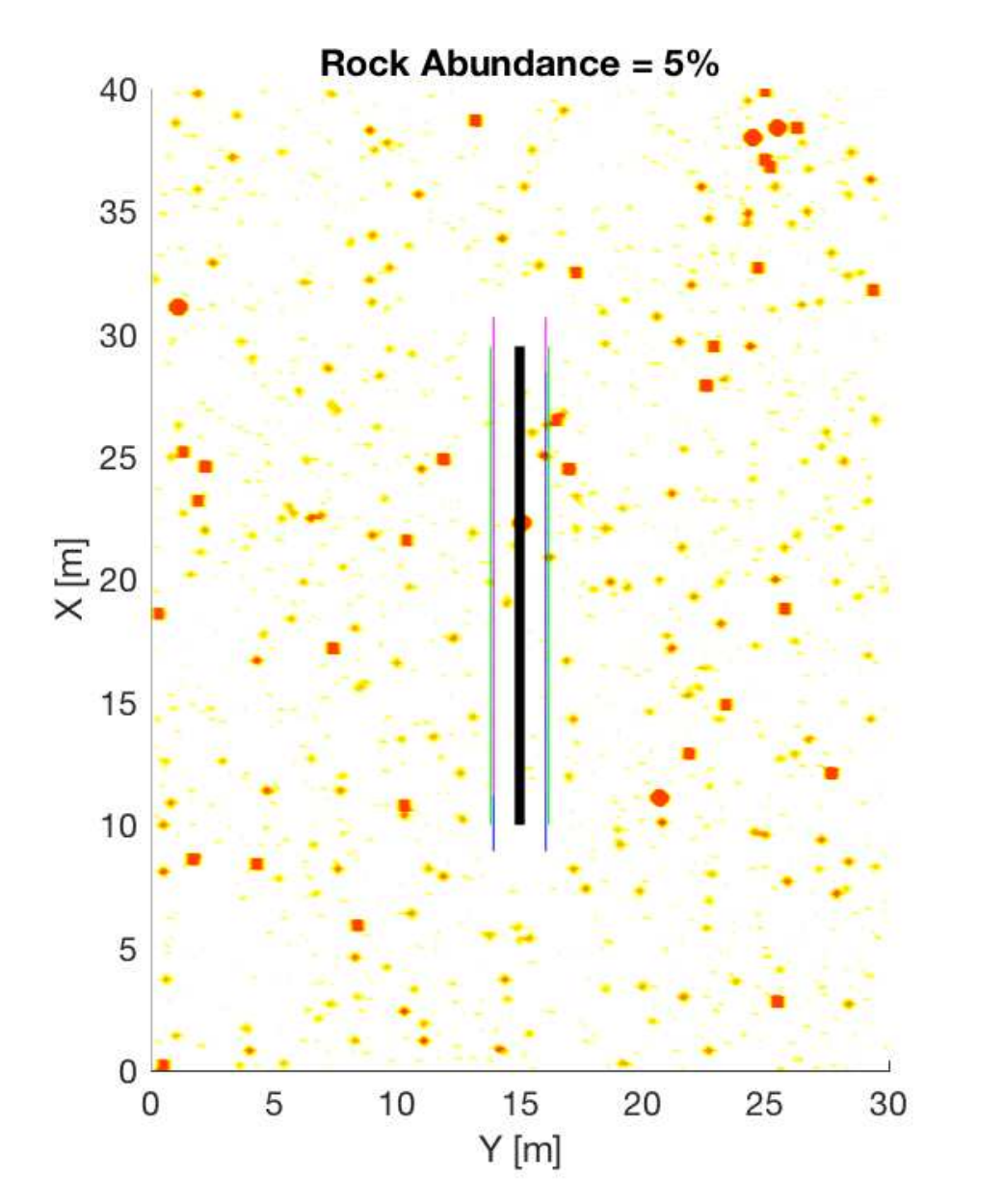}
        \includegraphics[width=0.24\textwidth]{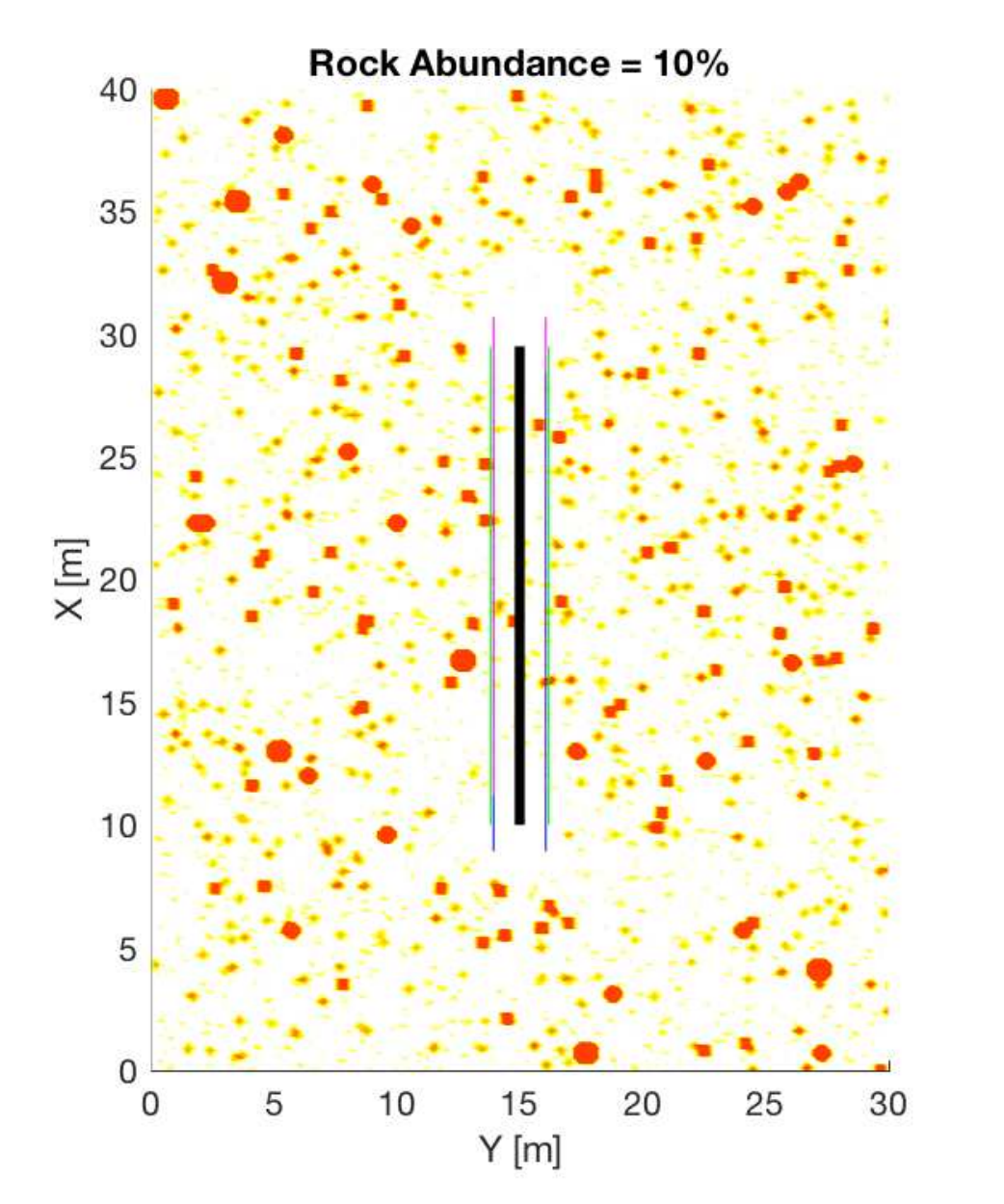}
        \includegraphics[width=0.24\textwidth]{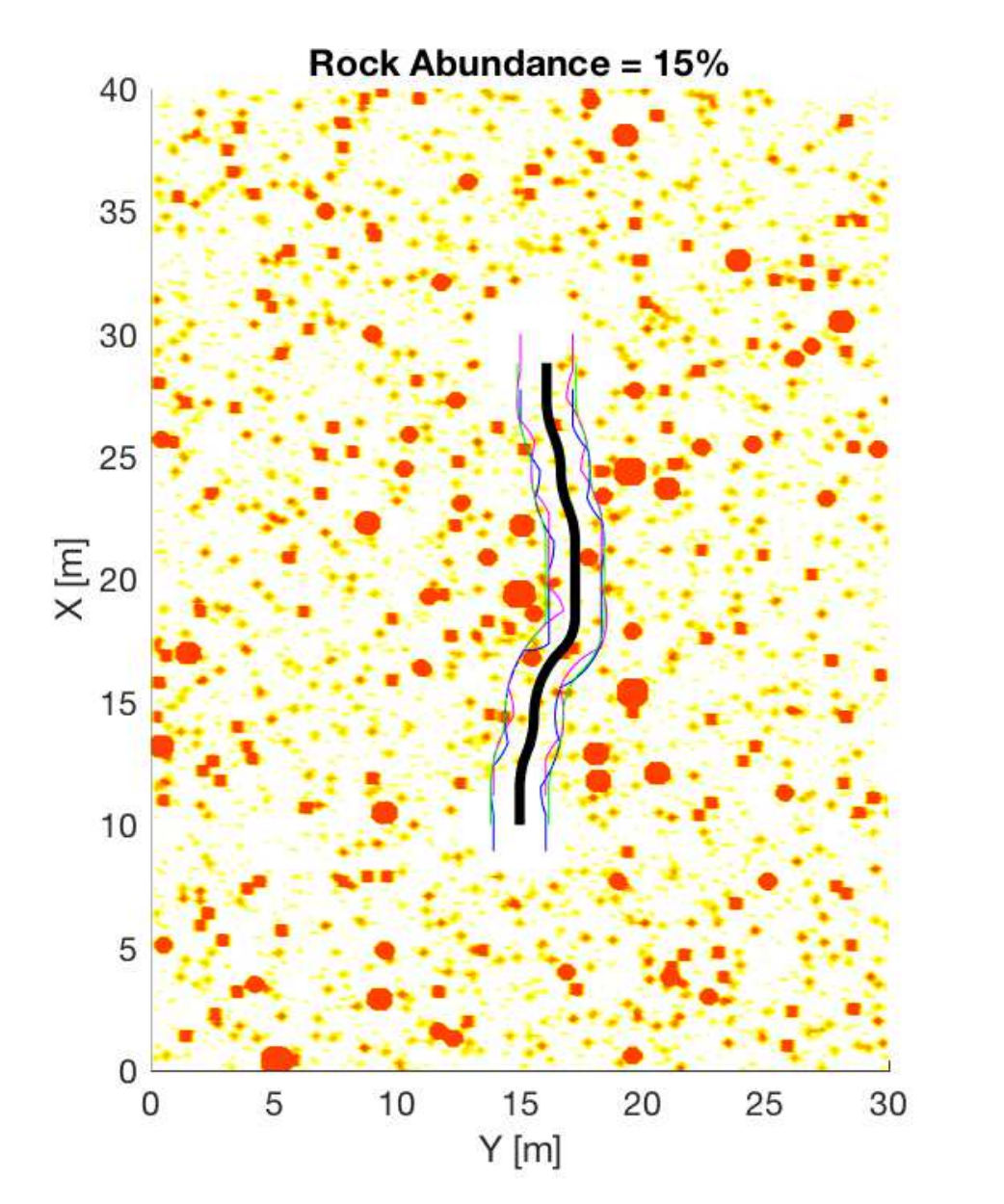}
        \includegraphics[width=0.24\textwidth]{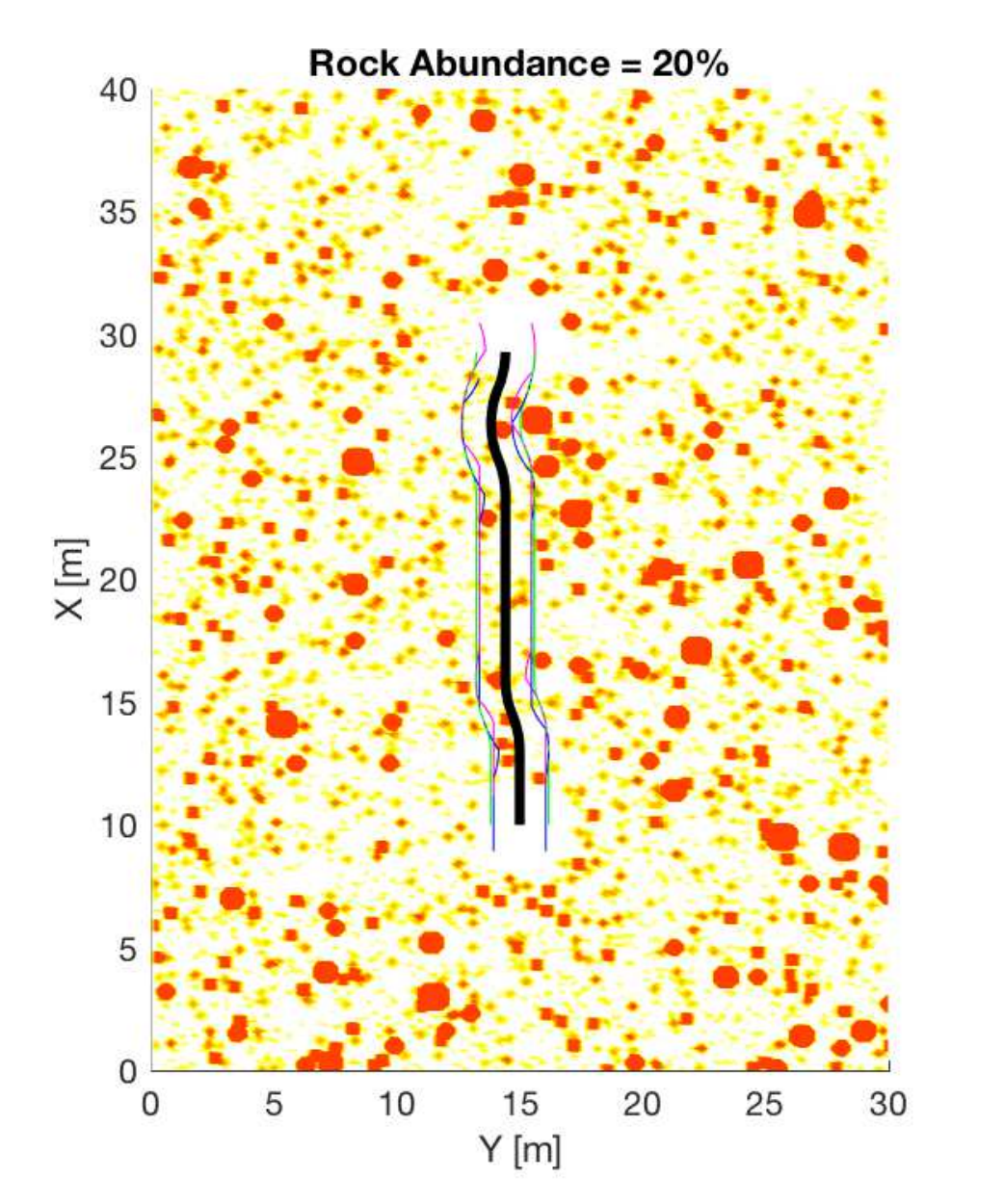}
        \caption{Ideal path planner}
    \end{subfigure}
    \caption{Comparison of safety assessment methods in 20\,[m] path planning with varying CFA levels. a) Conventional method that checks slope and step hazards with rover-sized inflation; b) Assessment with worst-case state from ACE bounds; c) Assessment with ground-truth state.}
    \label{fig:path_planning}
\end{figure}

\begin{figure}
    \centering
    \begin{subfigure}{0.48\textwidth}
        \includegraphics[width=0.9\textwidth]{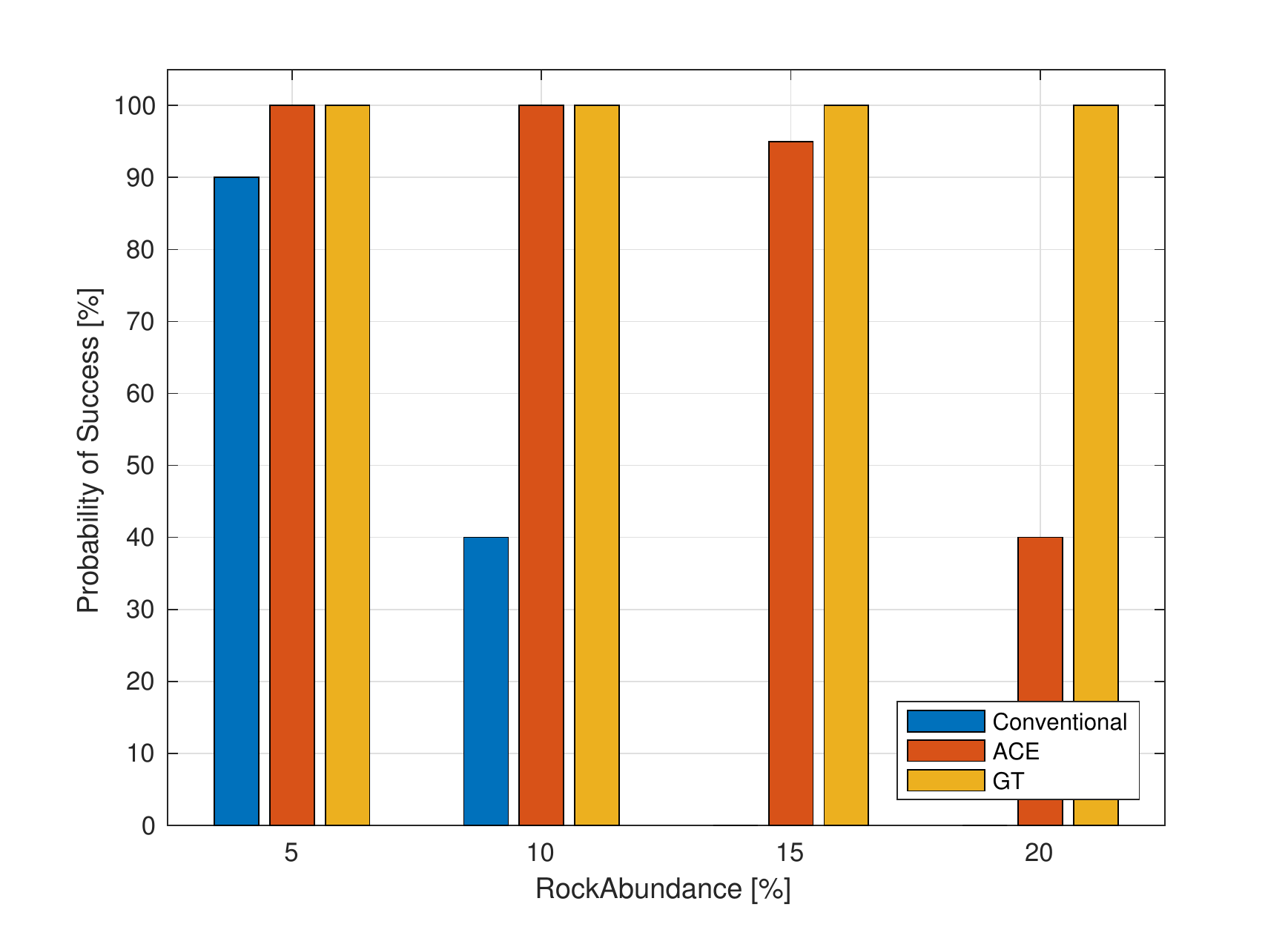}
        \caption{Probability of success}
    \end{subfigure}
    \begin{subfigure}{0.48\textwidth}
        \includegraphics[width=0.9\textwidth]{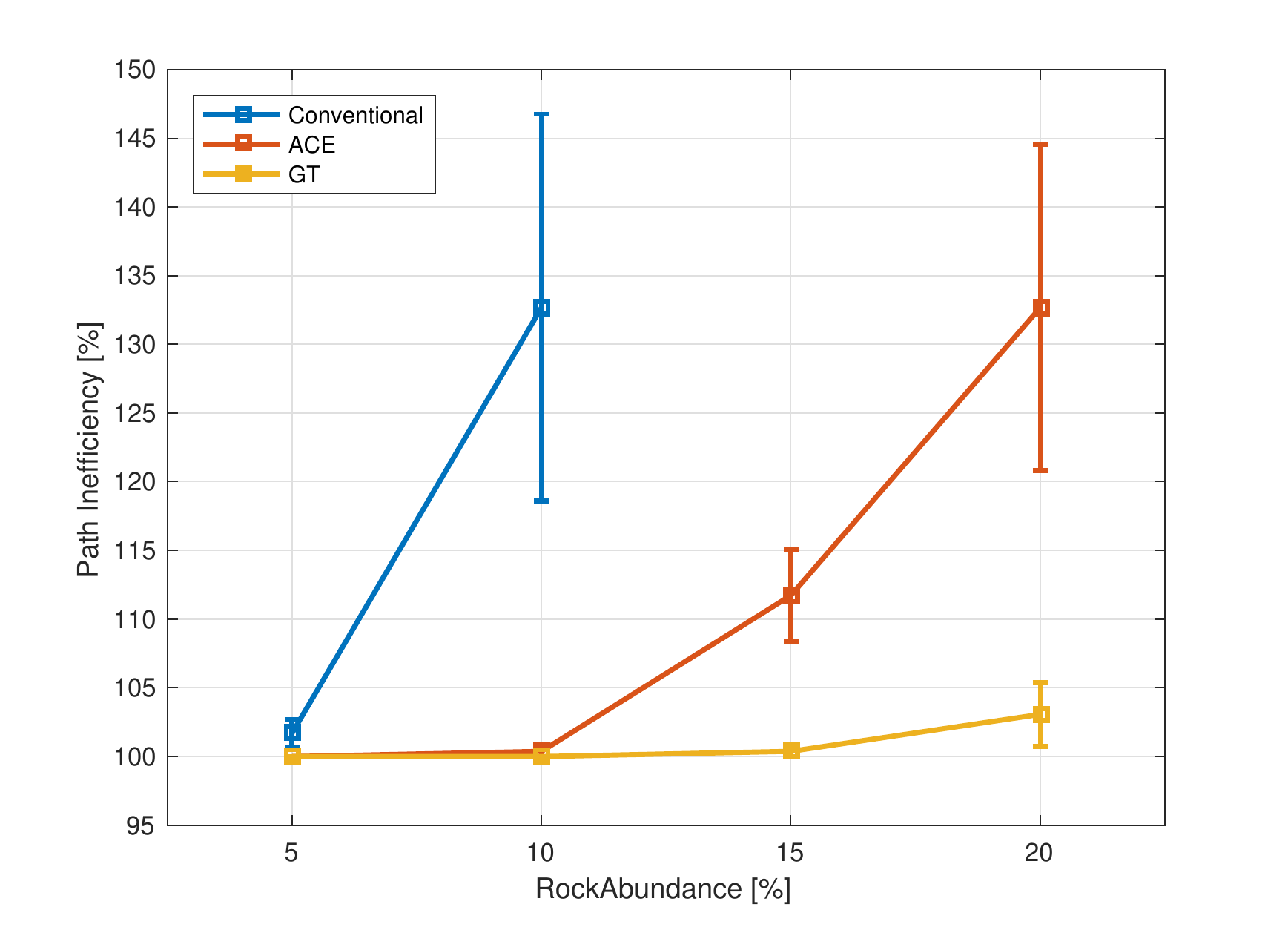}
        \caption{Average path inefficiency with standard error}
    \end{subfigure}
    \caption{Statistical result of path planning over 20 different maps in each CFA level. }
    \label{fig:path_planning_stats}
\end{figure}

Finally, we directly compared the performance of ACE-based path planning with the state-of-the-art in simulation.
The point of comparison was a variant of GESTALT implemented in MATLAB\footnote{We did not use the flight implementation of GESTALT because porting a part of spacecraft flight software is difficult due to technical and security reasons.}, which computes slope, roughness, and step hazards from plane fitting, and creates a goodness map by inflating hazards by the rover radius. 
In addition, we also compared against the ``ideal" path planner that uses the ground-truth collision check (no conservatism).
Such a planner is computationally unacceptable for the practical Mars rovers, but the comparison gives us an insight about how close the ACE-based paths are to the strictly optimal paths.
The path planning algorithm is the same for all the three planners; a depth-five tree search was used for path selection with 1.5\,[m] edge for each depth, while the collision check was run at every 0.25\,[m]. The only difference is the collision check method. 

The terrains we tested are flat, 30-by-40 meters in size, randomly populated with rocks at four different CFA levels (5, 10, 15, 20\%). 
We created 20 terrains for each CFA levels (80 terrains in total). 
Three planners were run on each of the 80 terrains.
A Curiosity-sized rover was commanded to go to the point 20\,[m] away. 
Two quantitative metrics were used for the comparison.
The first is the path inefficiency, defined as the fraction of the generated path length and the straight-line distance.
Intuitively, the over-conservatism of collision checking should result in an increased path inefficiency because it is more likely that the paths heading straightly towards the goal is incorrectly judged unsafe, resulting in a highly winding path. 
The second metric is the success rate, defined as the number of runs the planner successfully arrived in the goal divided by the total number of runs.
An excessive conservatism may result in a failure to reach the goal because no feasible path is found to move forward. 

\autoref{fig:path_planning} shows representative examples of paths generated by the three methods. 
The top, middle, and bottom rows are the state-of-the-art, ACE-based, and ideal path planners. 
As expected, the state-of-the-art paths were most winding (greater path inefficiency) while the ideal paths were the most straight. 
Notably, the state-of-the-art approach failed to find a path to the goal at 15 and 20\% \ac{CFA}, while the ACE-based planner were able to find a way to the goal.
The ACE-based planner was more capable of finding paths through cluttered environments mainly because it allows straddling over rocks if sufficient clearance is available. 
However, the ACE-based paths are less efficient compared to the ideal ones.
This result is again expected, because ACE conservatively approximates the rover states for the sake of significantly reduced computation (as reported in Section \ref{sec:runtime}) compared to the exact kinematic solution.

\autoref{fig:path_planning_stats} shows the statistical comparison over the 20 randomly generated maps for each CFA level in terms of the two quantitative metrics.  
According to \autoref{fig:path_planning_stats}(a), the ACE-based planner was capable of driving reliably ($\ge 95$\% success rate) up to 15\% CFA, but the success rate drops significantly at 20\% CFA.
In comparison, the state-of-the-art path planning had only 40 \% success rate at rather benign 10\% CFA terrains.
The ideal path planner was always be able to find a path to the goal for all the tested CFA values.
Next, the results on path inefficiency in \autoref{fig:path_planning_stats}(b) clearly shows the difference in algorithmic conservatism. 
For example, at 10\% CFA, the state-of-the-art planner resulted in 33\% path inefficiency while it was nearly zero for the ACE-based and the ideal planners. At 15\% CFA, the ACE-based planner resulted in 12\% path inefficiency while that of the ideal planner is still nearly zero. 
The path inefficiency of the state-of-the-art planner was not computed for 15 and 20\% CFA because the success rate was zero. 
Finally, at 20\% CFA, the path inefficiency of the ACE-based planner went up to 33\% while that of the ideal planner was at 3\%. 
The CFA of the landing site of the Mars 2020 Rover (Jezero Crater) is typically less than 15\%, while we can almost surely find a round to go around the fragmented spots with $\ge 15$\% CFA.
Therefore, with these results, ACE allows us to confidently drive the Mars 2020 rover autonomously for the majority of the drive.

\section{Conclusions} \label{sec:conclusion}

In this paper, we presented an approximate kinematics solver that can quickly, albeit conservatively, evaluate the state bounds of articulated suspension systems. The proposed method provides a tractable way of determining path safety with the limited computational resources available to planetary rovers. \ac{ACE} avoids expensive iterative operations by only solving for the worst-case rover-terrain configurations. The algorithm is validated using simulations and actual rover testbeds, giving satisfactory results in all experiments including 42 days of field test campaign.

The experimental results indicate that the ACE-based planner  successfully navigates the rover in environments with similar complexity to the planned landing site of Mars 2020 mission; however, one of the remaining algorithmic limitations is over-conservatism in estimated state bounds. Especially, the conservatism becomes greater on highly undulating terrain. An excessive conservatism may result in path inefficiency or a failure to find a path to the goal. Mitigating the extra conservatism is deferred to our future work.

Although the algorithm is primarily designed for planetary rover applications, the work is applicable to other domains where fast state estimation is needed but the fidelity of estimation is not demanded. Examples include trajectory planning of manipulators and path planning of ground/aerial/maritime vehicles. The importance of this method is in how we incorporate environmental uncertainty into the planning problem, without redundant computation or unsafe approximation. With the proper bounds of uncertainty, the robot state is guaranteed to be safe within well-defined intervals. 

The \ac{ACE} algorithm was successfully integrated with the surface navigation software of \ac{M2020} rover mission. \ac{ACE} will enable faster and safer autonomous traverse on more challenging terrains on the red planet.


\appendix

\section{Variable Definitions}

The following table introduces a list of variables used in this paper.

\begin{table}[h!]
\centering
\begin{tabular}{rl}
    \toprule
    Variable & Definition \\
    \midrule
    $(x,y,z,\phi,\theta,\psi)$ & 6 DoF pose \\
    $l_{df}$ & Link length between differential joint and front wheel \\
    $l_{dr}$ & Link length between differential joint and rear wheel (Rocker) \\
    $l_{bm}$ & Link length between bogie joint and middle wheel (Rocker-bogie) \\
    $l_{br}$ & Link length between bogie joint and rear wheel (Rocker-bogie) \\
    $z_{\{f,m,r\}}$ & Height of front, middle, and rear wheels \\
    $z_{d}$ & Height of differential joint\\
    $z_{b}$ & Height of bogie joint  \\
    $\delta_{\{l,r\}}$ & Angle change of left and right differential joints ($\delta_l=\delta_r$) \\
    $\beta_{\{l,r\}}$ & Angle change of left and right bogie joints \\
    $(x_{od}, y_{od}, z_{od})$ & Translational offset from the body frame origin to differential joint \\
    $z_o$ & Height of body frame origin \\
    $\kappa_{d0}$ & Angle between horizontal line and differential-front link on flat plane \\
    $\kappa_{b0}$ & Angle between horizontal line and bogie-middle link on flat plane \\
    \bottomrule
\end{tabular}
\end{table}

\section{Video Attachment}
The supplement movie visually presents the state bound propagation process from terrain heights to vehicle's attitude through rocker-bogie suspensions.


\subsubsection*{Acknowledgments}
This research was carried out at the Jet Propulsion Laboratory, California Institute of Technology, under a contract with the National Aeronautics and Space Administration. Copyright 2019 California Institute of Technology. Government sponsorship acknowledged.

\bibliographystyle{apalike}
\bibliography{ROB-18-0189.R2.bib}

\end{document}